\documentclass[lettersize,journal]{IEEEtran}

\usepackage[colorlinks,
            linkcolor=black,
            anchorcolor=black,
            citecolor=black]{hyperref} %

\usepackage{amsmath,amsfonts}
\usepackage{algorithm}
\usepackage{array}
\usepackage{textcomp}
\usepackage{stfloats}
\usepackage{url}
\usepackage{verbatim}
\usepackage{graphicx}
\usepackage{hyperref}

\usepackage{cleveref}
\crefname{figure}{Fig.}{Figs.}  
\crefname{table}{Tab.}{Tabs.}
\usepackage{cite}
\usepackage{subcaption}
\usepackage[table]{xcolor}
\usepackage{xspace}
\usepackage{booktabs}
\usepackage{multirow}
\usepackage{algpseudocode}
\usepackage[switch]{lineno}
\usepackage{hyperref}

\newcommand{\ie}{i.e.\@\xspace}
\newcommand{\eg}{e.g.\@\xspace}

\usepackage{xcolor}

\begin{document}

\title{SegTrans: Transferable Adversarial Examples for Segmentation Models}


\author{Yufei Song, Ziqi Zhou, Qi Lu, Hangtao Zhang, Yifan  Hu, Lulu Xue,  Shengshan Hu,~\IEEEmembership{Member,~IEEE,} Minghui Li,~\IEEEmembership{Member,~IEEE,} Leo Yu Zhang,~\IEEEmembership{Member,~IEEE}


\thanks{Correspondence to Dr Ziqi Zhou.
Yufei Song, Qi Lu, Yifan Hu, Hangtao Zhang, Lulu Xue, and Shengshan Hu are with the School of Cyber Science and Engineering, Huazhong University of Science and Technology, Wuhan, Hubei, China 
(e-mail:
~\href{mailto:yufei17@hust.edu.cn}{yufei17@hust.edu.cn}; 
~\href{mailto:luqi@hust.edu.cn}{luqi@hust.edu.cn};
~\href{mailto:hyf1009@hust.edu.cn}{hyf1009@hust.edu.cn};
~\href{mailto:zhanghangtao7@163.com}{hangt\_zhang@hust.edu.cn};
~\href{mailto:lluxue@hust.edu.cn}{lluxue@hust.edu.cn};
~\href{mailto:hushengshan@hust.edu.cn}{hushengshan@hust.edu.cn}).

Ziqi Zhou is with the School of Computer Science and Technology, Huazhong University of Science and Technology, Wuhan, Hubei, China (e-mail:~\href{mailto:zhouziqi@hust.edu.cn}{zhouziqi@hust.edu.cn}).


Minghui Li is with the School of Software Engineering, Huazhong University of Science and Technology, Wuhan, Hubei, China (e-mail:~\href{mailto:minghuili@hust.edu.cn}{minghuili@hust.edu.cn}).

Leo Yu Zhang is with the School of Information
and Communication Technology, Griffith University, Southport, Queensland,
Australia (e-mail: \href{mailto:leo.zhang@griffith.edu.au}{leo.zhang@griffith.edu.au}).

}


}

\markboth{IEEE TRANSACTIONS ON MULTTIMEDIA,~Vol.xxx}%
{Shell \MakeLowercase{\textit{et al.}}: A Sample Article Using IEEEtran.cls for IEEE Journals}



\maketitle
\begin{abstract}
Segmentation models exhibit significant vulnerability to adversarial examples in white-box settings, but existing adversarial attack methods often show poor transferability across different segmentation models.
While some researchers have explored transfer-based adversarial attack (\ie, transfer attack) methods for segmentation models, the complex contextual dependencies within these models and the feature distribution gaps between surrogate and target models result in unsatisfactory transfer success rates. 
To address these issues, we propose SegTrans, a novel transfer attack framework that divides the input sample into multiple local regions and remaps their semantic information to generate diverse enhanced samples. These enhanced samples replace the original ones for perturbation optimization, thereby improving the transferability of adversarial examples across different segmentation models. 
Unlike existing methods, SegTrans only retains local semantic information from the original input, rather than using global semantic information to optimize perturbations.
Extensive experiments on two benchmark datasets, PASCAL VOC and Cityscapes, four different segmentation models, and three backbone networks show that SegTrans significantly improves adversarial transfer success rates without introducing additional computational overhead. 
Compared to the current state-of-the-art methods, SegTrans achieves an average increase of 8.55\% in transfer attack success rate and improves computational efficiency by more than 100\%.
The code is available at \url{https://github.com/Yufei-17/SegTrans}.
\end{abstract}

\begin{IEEEkeywords}
Semantic Segmentation, Transfer Attack
\end{IEEEkeywords}
 
\section{Introduction}\label{sec:inroduction}
With the development of deep learning, semantic segmentation models are playing an increasingly important role in complex scenarios such as autonomous driving~\cite{xiao2023baseg}, medical image analysis~\cite{kalinin2020medical}, and remote sensing~\cite{du2021incorporating}.
By performing pixel-level classification, these models~\cite{zhao2017pyramid,chen2014semantic} achieve accurate object segmentation.
However, existing studies demonstrate that segmentation models are vulnerable to  adversarial examples~\cite{gu2022segpgd,jia2023transegpgd,song2025segment, song2025seg, wang2025breaking, wang2025advedm, advclip, zhou2023downstream, zhou2025sam2} and poisoning attacks~\cite{zhang2024detector, wang2024trojanrobot,zhang2025test,wan2025mars,wang2024unlearnable,li2025detecting, wang2024eclipse,zhou2025darkhash,yu2025spa}. 
Among them, adversarial examples are more practical for real-world deployment, as their core concept is to add imperceptible perturbations to the input data, thereby misleading deep learning models into producing incorrect outputs.
Specifically, in autonomous driving, prior studies show that attackers use adversarial stickers to mislead the recognition of traffic signs and the detection of pedestrians, severely compromising driving safety~\cite{nesti2022evaluating}.

\begin{figure}[t]
 \setlength{\abovecaptionskip}{4pt}
    \centering
    \includegraphics[scale=0.33]{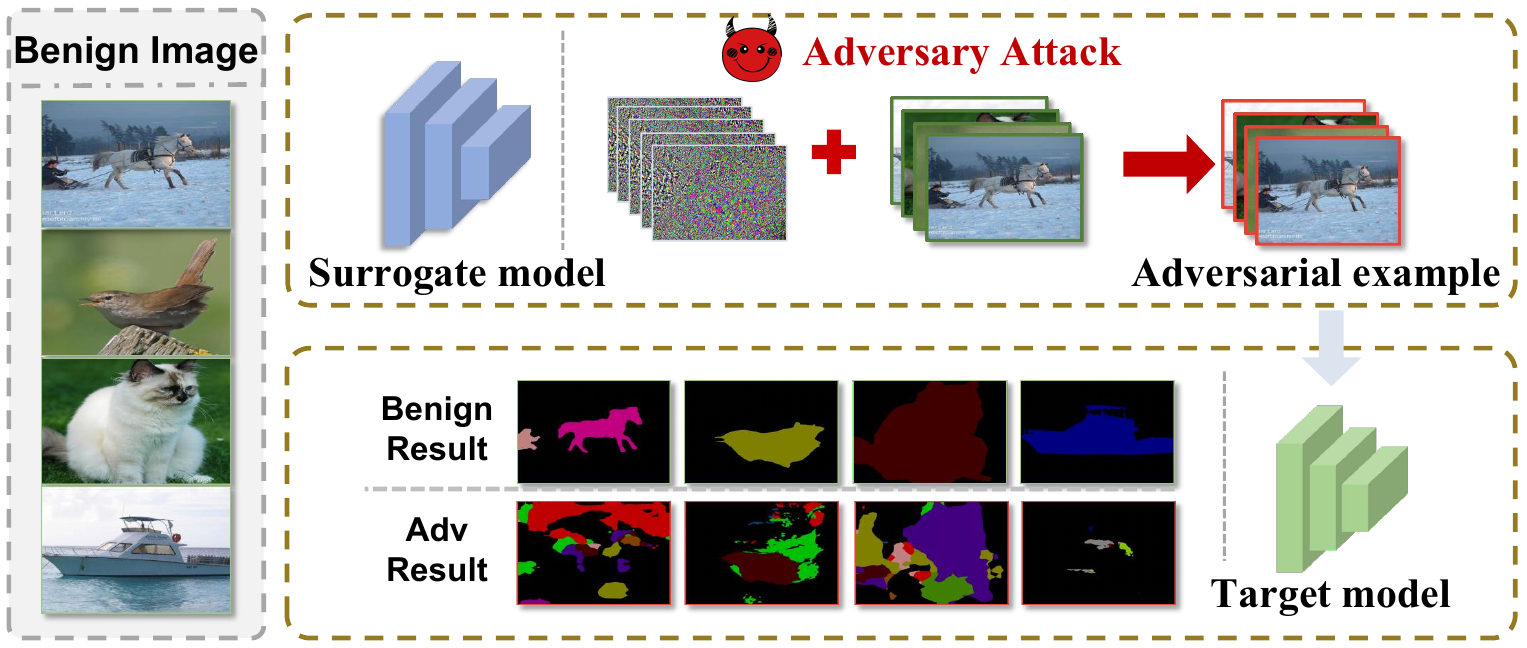}
    \caption{
    Illustration of transfer-based adversarial attacks
    }
    \label{fig:demo}
  \vspace{-0.4cm}
\end{figure}

Recent studies~\cite{jia2023transegpgd,cai2023ensemble} have begun exploring transfer-based adversarial attacks (\ie, transfer attacks) against segmentation models, which do not require access to any information about the target model. Instead, attackers utilize a surrogate model to generate transferable perturbations that can fool the target model~\cite{narodytska2017simple}, as illustrated in \cref{fig:demo}.
For instance, TranSegPGD~\cite{jia2023transegpgd} improves the generalization of perturbations by dynamically assigning weights to pixels with varying attributes during training. EBAD~\cite{cai2023ensemble} adopts a model-ensemble strategy, leveraging multiple surrogate models simultaneously to strengthen adversarial examples.
However, these methods overlook the strong semantic contextual dependencies between objects in segmentation tasks and fail to address the over-reliance of adversarial examples on surrogate model features, particularly under limited computational budgets. To the best of our knowledge, designing an effective transfer attack for segmentation models without incurring additional computational overhead remains a significant and unresolved challenge.

In our view, designing an effective transfer attack against segmentation models poses two major challenges.
First, segmentation models leverage semantic context to correct misclassified target objects, and this tight coupling phenomenon significantly reduces the effectiveness of adversarial examples~\cite{chen2017deeplab}. Second, the complex architectures of these models lead to perturbations being overly dependent on fixed patterns of the surrogate model~\cite{rice2020overfitting}, thereby limiting their transferability.
This raises a critical question:

\begin{quote}
    \emph{Is it possible to enhance the transferability of adversarial examples by only leveraging the inherent characteristics of segmentation models?}  
\end{quote}

To address this challenge, we propose SegTrans, a novel transfer attack framework aimed at improving the transferability of adversarial examples in segmentation tasks. Unlike classification models that primarily rely on global features, segmentation models emphasize contextual relationships between objects in an image~\cite{garcia2017review} (\eg, a bicycle ridden by a person is segmented more accurately than one without a rider).
Therefore, our intuition is to disrupt the connections between different objects in the image to achieve a successful attack.

To this end, the SegTrans framework is composed of two key modules: the multi-region perturbation activation module and the semantic remapping module. The former generates enhanced samples by retaining partial semantic information from the original input, while the latter refines the perturbation by substituting the original sample with multiple enhanced variants.  
Specifically, in the first module, the image is divided into multiple identical rectangular regions, with only partial semantic information preserved in each region. In the second module, to mitigate the perturbation’s reliance on the surrogate model, multiple enhanced versions of the same sample are generated and utilized to construct the feature representation of the attack target within SegTrans.

By shifting from attacks based on global semantic information to those leveraging partial semantics, we substantially improve the transferability of adversarial examples across diverse segmentation models.
Experiments on four segmentation models, including FCN~\cite{long2015fully}, PSPNet~\cite{zhao2017pyramid}, DeepLabV1~\cite{chen2014semantic}, and DeepLabV3+~\cite{xing2020encoder}, as well as on two benchmark segmentation datasets, demonstrate that SegTrans achieves a high attack success rate.

Our main contributions are summarized as follows:
\begin{itemize}
\item 
We conduct an in-depth investigation into the tight coupling phenomenon in segmentation tasks, discovering its role in enhancing the transferability of adversarial attacks, and further revealing the vulnerabilities of segmentation models.
\item 
We propose a novel transfer attack framework that consists of two modules, multi-region perturbation activation and semantic remapping, which transform attacks utilizing global semantic information into attacks leveraging partial semantic information.
\item 
The qualitative and quantitative experimental results across four segmentation models and two benchmark segmentation datasets show that SegTrans achieves a high attack success rate and strong transferability.
\end{itemize}
\section{Related Works}

\noindent\textbf{Segmentation models.} Segmentation models typically incorporate architectures such as Image Pyramid~\cite{wu2022fpanet}, Encoder-Decoder~\cite{xing2020encoder}, Context Module~\cite{wu2020cgnet}, Spatial Pyramid Pooling~\cite{ru2023forest}, Atrous Convolution (AC)~\cite{chen2014semantic}, and Atrous Spatial Pyramid Pooling (ASPP)~\cite{chen2017deeplab}.
These architectures enhance the model's ability to capture image details by introducing multi-scale feature extraction, strengthening contextual information, or expanding the receptive field, resulting in more precise segmentation outcomes.
Specifically, DeepLabV1~\cite{chen2014semantic} and PSPNet~\cite{zhao2017pyramid} use the AC structure to enhance global context awareness while preserving local details without sacrificing resolution.
After DeeplabV2~\cite{chen2017deeplab}, subsequent versions~\cite{chen2017rethinking,chen2018encoder} adopted the ASPP structure, which enhances the model's understanding of complex scenes and object boundaries through multiscale feature aggregation. 
Different models have different architectures, which makes it difficult for adversarial samples generated by surrogate models to transfer to target models.

\noindent\textbf{Adversarial examples.} 
Existing white-box adversarial attacks~\cite{zhou2025numbod, zhou2024securely, zhou2024darksam} against segmentation models have achieved promising performance. 
SegPGD~\cite{gu2022segpgd} improves attack success by dynamically adjusting the learning rate and the weight of the attacked regions.
CosPGD~\cite{agnihotri2024cospgd} uses a simple alignment score to smoothly and differentiably adjust the loss function.
Despite their strong performance in white-box settings, these methods generally exhibit weak transferability.
To enhance the generalization of adversarial examples across different segmentation models, recent studies have focused on transfer-based adversarial attacks.  
TranSegPGD~\cite{jia2023transegpgd} enhances SegPGD by incorporating KL divergence~\cite{kullback1951information} into pixel selection, boosting adversarial example transferability across segmentation models. However, differences in architecture, context modeling, and feature extraction among models limit its effectiveness.  
EBAD~\cite{cai2023ensemble} employs an ensemble strategy to dynamically weight surrogate models, generating transferable adversarial examples without target model access. Despite improving transferability, its high computational cost reduces practicality.

\noindent\textbf{Adversarial defense.} 
To counter adversarial attacks, several defense methods have been proposed.
Adversarial training~\cite{xing2021algorithmic} enhances model robustness by incorporating adversarial examples into the training process, enabling segmentation models to better resist perturbations~\cite{lupart2023study}.
Model pruning~\cite{xiang2024adapmtl,lu2024generic} simplifies the model by removing redundant weights or neurons, eliminating vulnerable connections.
Corruption~\cite{guo2018countering} introduces noise or perturbations to adversarial examples, reducing their impact and improving the model's robustness.
These popular methods will be used to validate the robustness of SegTrans against defenses.

\section{Methodology}
\label{sec:methodology}
\subsection{Problem Formulation}
\label{sec:problem}
We assume the attacker has no access to any information about the target segmentation model and training data.
The attacker can construct a surrogate model that closely resembles the target model and use it to train adversarial perturbations on a dataset $\mathcal{D}$. 
Their goal is to design adversarial perturbations \( \delta \) that cause the target model $f_{\text{target}}$ to produce incorrect segmentation results when processing the input image, thus degrading the model's performance.
The perturbation \( \delta \) is constrained by an \( \ell_p \)-norm with a budget \( \epsilon \) and is required to remain imperceptible.
This optimization problem can be formulated as:

\begin{equation}
\begin{array}{c}
    \max_{\delta} \mathbb{E}_{x \sim \mathcal{D}} \left[ f_{\text{target}}(x + \delta) \neq f_{\text{target}}(x) \right] \\
    \text{s.t.} \quad \| \delta \|_p \leq \epsilon
\end{array}
\label{eq:problem}
\end{equation}

\subsection{Intuition behind SegTrans}
\label{sec:intuitions}
Unlike classification models that assign a single label to the entire image, segmentation models predict a label for each pixel.
This pixel-wise prediction makes transfer attacks significantly more challenging in segmentation tasks, as the adversarial perturbation must not only disrupt global image features (the complete semantic characteristics of the entire image) but also effectively compromise local image features (the local semantic characteristics of objects within the image) and pixel-level representations to degrade segmentation performance.
Specifically, we identify two key challenges in this setting.

\begin{figure}[!t]
  \setlength{\abovecaptionskip}{4pt}
    \centering
    \includegraphics[scale=0.5]{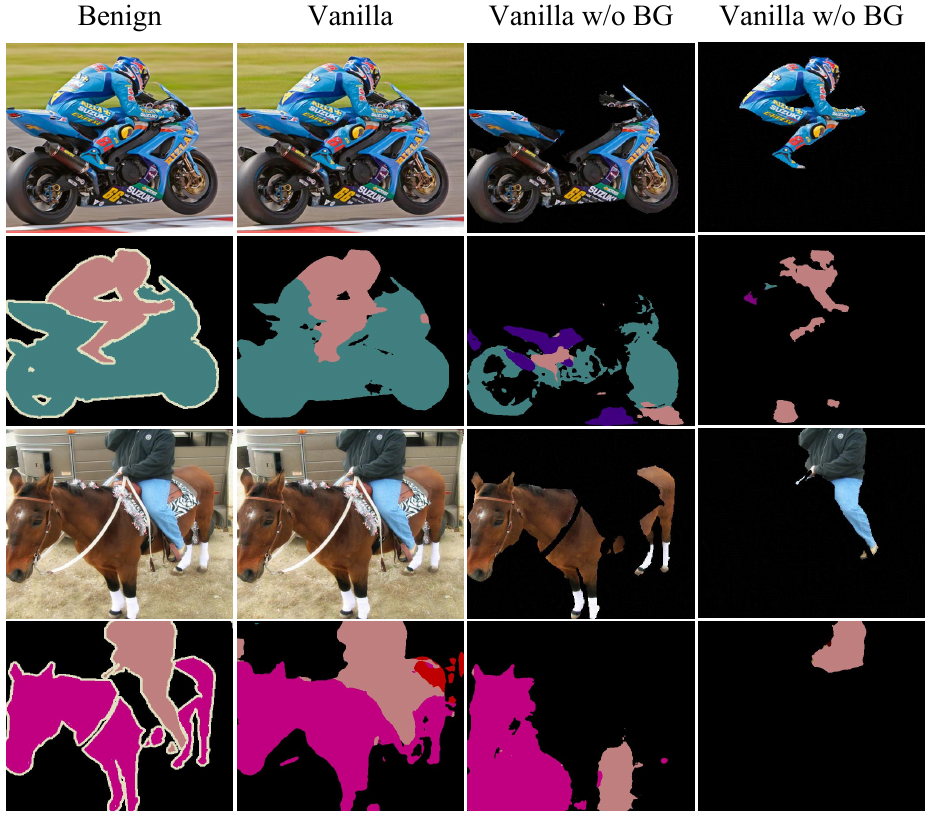}

    \caption{
    The perturbation failure caused by tight coupling phenomenon between objects.
    Vanilla and Vanilla w/o BG represent the attack results of vanilla on the original image and the attack results of vanilla after removing the background outside the target, respectively. 
    Here, "Vanilla" refers to the baseline attack method PGD~\cite{madry2018towards} that we have selected.
    }

    \label{fig:tight}
     \vspace{-0.4cm}
\end{figure}

\noindent\textbf{Challenge I: The perturbation failure caused by tight coupling phenomenon.} 
In semantic segmentation tasks, objects often exhibit strong semantic associations, such as the relationship between a motorcycle and a person riding it~\cite{garcia2017review}. 
These associations enable the model to leverage contextual information for precise segmentation, while mitigating the impact of adversarial perturbations.
We refer to this semantic relationship as the \textit{tight coupling phenomenon} between objects.
Specifically, the tight coupling phenomenon enables the model to rely not only on features of individual objects during prediction but also to incorporate contextual information by exploiting relationships between objects to infer correct segmentation boundaries. 
Upon application of adversarial perturbations to an object, the model adjusts its predictions through this phenomenon, thereby effectively mitigating the impact of such perturbations (see \cref{fig:tight}).
We believe this phenomenon mainly arises from the dense prediction nature of semantic segmentation and its strong reliance on spatial context. 
Segmentation models make classification decisions for each pixel, which causes them to depend heavily on semantic relationships between pixels and the overall context, leading to tight coupling between features. 
In contrast to segmentation models, image classification models output a single label for the whole image, so spatial context has less influence, and the tight coupling phenomenon is less apparent.
Moreover, the multi-scale feature fusion and more complex context modeling in segmentation models further increase feature dependencies, making perturbations harder to transfer effectively during attacks.

To address this challenge, we propose a Multi-region Perturbation Activation Strategy.
In each epoch, the method randomly selects multiple non-overlapping regions as attack targets, disrupting the contextual integrity of the input sample.
Instead of relying on complete semantic content, the approach enhances the sample by retaining several randomly selected, non-overlapping rectangular regions and uses these enhanced samples to guide perturbation optimization.
This strategy effectively disrupts the input’s semantic information, mitigates the smoothing effect induced by the tight coupling phenomenon, and enhances the transferability of adversarial examples across models through generating more diverse training samples.

\begin{figure}[!t]
  \setlength{\abovecaptionskip}{4pt}
    \centering
    \includegraphics[scale=0.58]{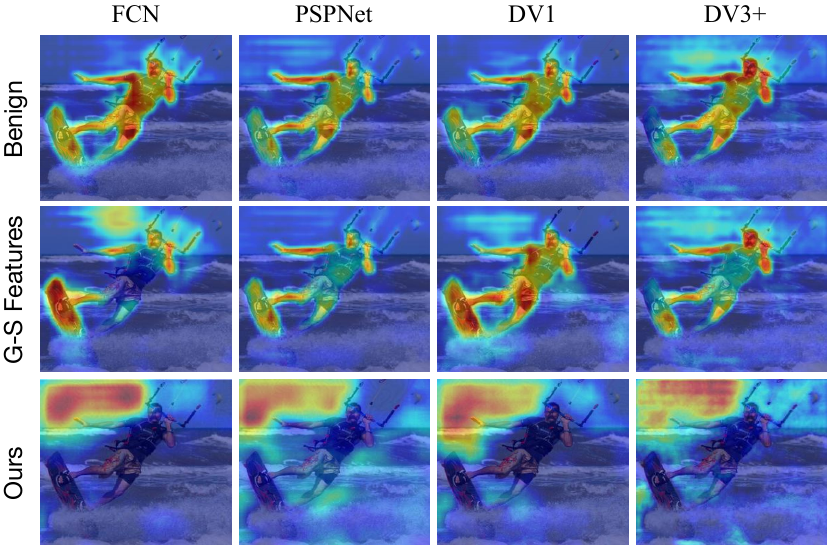}

    \caption{Grad-CAM visualizations of different models on the same input image, where each row represents the same sample.
    FCN serves as the surrogate model, while the other three are target models.
    Global semantic features, DeepLabV1 and DeepLabV3+ are abbreviated as G-S features, DV1 and DV3+, respectively.}
    \label{fig:gradcam}
     \vspace{-0.4cm}
\end{figure}

 \begin{figure*}[!t]
    \centering
    \includegraphics[scale=0.55]{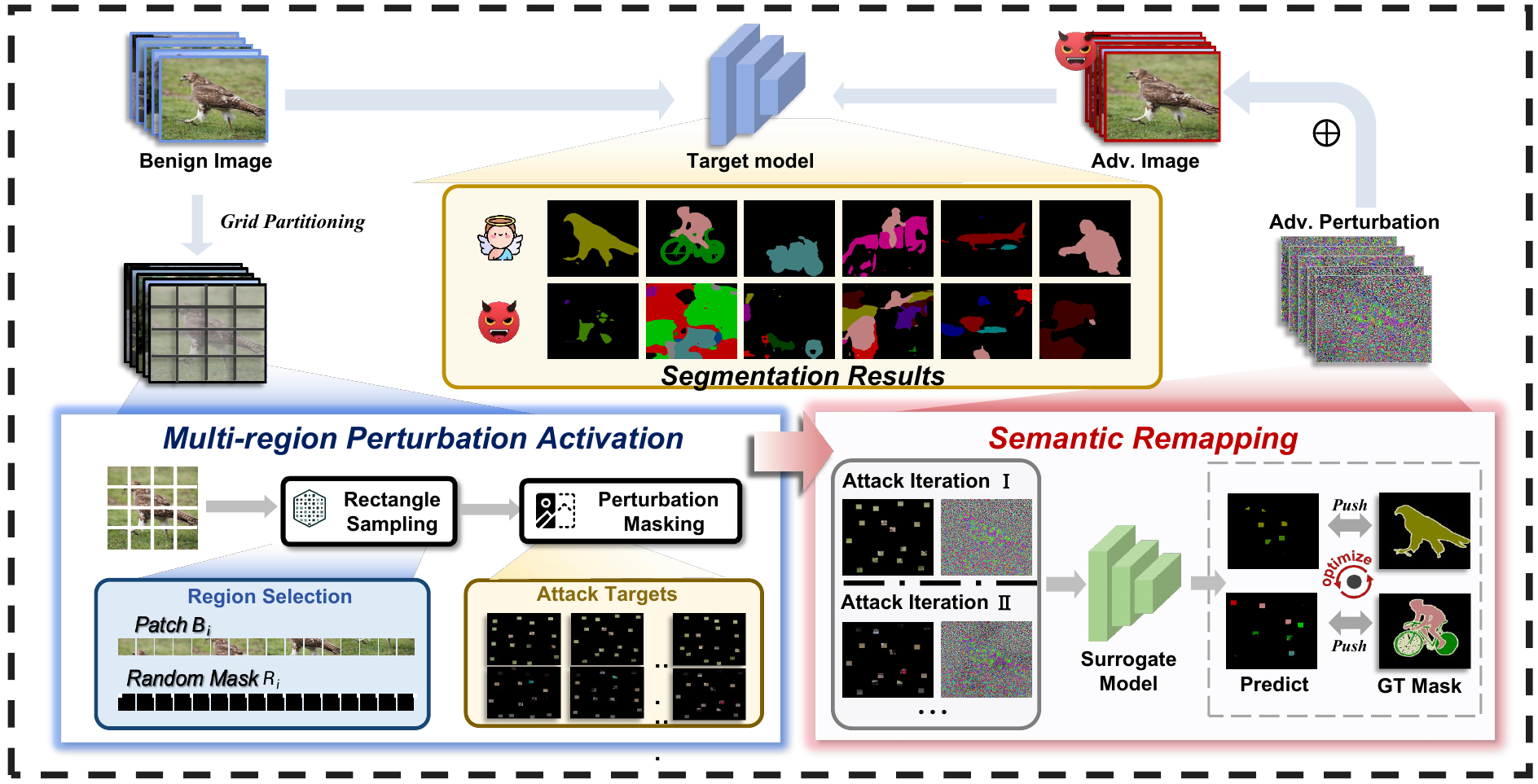}
    \caption{The framework of SegTrans}
    \label{fig:pipeline}
      \vspace{-0.4cm}
\end{figure*}

\noindent\textbf{Challenge II: The limited effectiveness of transfer attacks induced by feature fixation.}
In transfer-based attacks, the “feature fixation” phenomenon refers to the tendency of adversarial perturbations to overfit the feature distribution of the surrogate model, resulting in poor transferability to target models with different architectures. 
This happens because segmentation models like FCN~\cite{long2015fully}, PSPNet~\cite{zhao2017pyramid}, DeepLabV1~\cite{chen2014semantic}, and DeepLabV3+~\cite{xing2020encoder} differ significantly in receptive field size, context modeling, and attention mechanisms. As a result, they focus on different regions of the same input image.
To better understand the phenomenon of feature fixation, we use the Grad\text{-}CAM~\cite{selvaraju2017grad} visualization method to reveal differences in feature attention regions for the same adversarial sample across different models. 
Grad-CAM computes the gradients of the input image and generates heatmaps, where different colors reflect the model’s varying attention to different regions of the sample.
The top row of \cref{fig:gradcam} presents the Grad-CAM visualization results of different models on the same input image.
It can be observed that different models show significant differences in the feature attention regions (e.g., the red areas) of the target object in the image, indicating that each model focuses on different aspects of the same input~\cite{long2022frequency}. 
This observation suggests that adversarial perturbations crafted for a specific model are closely tied to its feature distribution, which limits their transferability and significantly reduces their effectiveness on other models.

Given that feature fixation arises from the model's excessive reliance on specific features of the input data, we propose a Semantic Remapping Strategy to reduce its impact by simulating the differences in how various models process images. 
Specifically, instead of using the original sample alone, we guide perturbation optimization with multiple semantically enhanced versions of the same input, thereby promoting diversity.
As illustrated in the second and third rows of \cref{fig:gradcam}, compared to attacks that rely solely on global semantics from the original sample, our strategy substantially reduces the feature representation bias between the surrogate and target models, effectively mitigating the adverse effects of feature fixation.

\subsection{SegTrans: A Complete Illustration}
To overcome the limitations of the tight coupling phenomenon and feature fixation in transfer attacks for segmentation tasks, we propose SegTrans. 
Specifically, SegTrans consists of two modules, namely Multi-region Perturbation Activation and Semantic Remapping.
The former aims to disrupt the integrity of semantic information in the input sample to reduce the impact of the tight coupling phenomenon on attack performance, while the latter mitigates the issue of feature fixation by optimizing the feature representations of the input sample.
We provide the detailed optimization process of SegTrans in \cref{alg:SegTrans}. 

\noindent\textbf{Multi-region perturbation activation.}
Existing transfer attack methods typically optimize perturbations based on the global semantic features of the sample, which causes the perturbations to be diminished by the target model's understanding of the contextual semantic information within the image.
To solve this problem, SegTrans preserves part of the semantic information from the original sample to obtain enhanced samples.
These regions are distributed globally and isolated locally, thereby disrupting the tight coupling phenomenon between the targets. Specifically, this strategy divides the given input image $x \in \mathbb{R}^{w \times h \times c}$ into $N$ identical grids.
Within each grid, a rectangle is randomly generated, where the length $l_{i}$ and width $w_{i}$ of the rectangle satisfy the following constraints:
\begin{equation}
l_i \in [0, G_w], \quad w_i \in [0, G_h], \quad \frac{l_i \times w_i}{G_w \times G_h} = \alpha_i
\label{eq:2}
\end{equation}
In the entire image sample, there are $N$ such random rectangular regions $\{R_1, R_2, \ldots, R_N\}$, which collectively satisfy the following constraint:
\begin{equation}
\sum_{i=1}^{N}(l_i \times w_i) = \sum_{i=1}^{N}(\alpha_i \times G_w \times G_h)
\label{eq:3}
\end{equation}
where $G_w$ and $G_h$ represent the width and height of each grid, respectively. $l_i$ and $w_i$ denote the length and width of each rectangular region, while $\alpha_i$ is a random number representing the ratio of the area of rectangle $R_i$ to the area of its corresponding grid.

\noindent\textbf{Semantic remapping.}
This is another key component of SegTrans, aimed at increasing the diversity of perturbations and reducing the impact of feature fixation.
Existing transfer attacks on segmentation models use the global semantic features of the sample to guide the optimization of perturbations, leading to perturbations that excessively depend on the fixed global features of the surrogate model.
As a result, these adversarial examples often exhibit reduced effectiveness when applied to other models.
To address this, Semantic Remapping divides each attack iteration into multiple attacks guided by the semantic features of random regions.
Specifically, in each attack iteration, we select $T$ batches of $N$ random rectangular regions from the input image, resulting in $T$ enhanced images that replace the original image.
These enhanced images gradually reconstruct the sample's semantic information over multiple optimization steps. 
This approach effectively increases the diversity of perturbations, significantly enhancing the transferability of the attack.
The constraint can be described as follows:
\begin{equation}
\sum_{i=1}^{T \times N}(l_i \times w_i) = \sum_{i=1}^{T \times N}(\alpha_i \times G_w \times G_h)
\label{eq:4}
\end{equation}

The attack process of SegTrans is guided by the overall optimization objective $\mathcal{L}_{adv}$, which aims to maximize the effect of adversarial perturbations by computing gradients over multiple regions of the input image. 
The optimization is performed using the following objective:

\begin{equation}
\mathcal{L}_{adv} = -\sum_{j=1}^{T} \mathcal{L}_{CE}\left( f_{surrogate}\left( (x + \delta) \odot \mathbf{m}_j \right), y \right)
\label{eq:5}
\end{equation}
where $\mathcal{L}_{CE}$ denotes the cross-entropy loss, $\delta$ is the perturbation added to the input $x$, and the regions of the image affected by the perturbation are determined by the mask $\mathbf{m}$.
Our design allows overlaps between the regions selected in different batches (indexed by $j$).
All these perturbations from different batches are then accumulated into the final total perturbation $\delta$.
The mask is generated by randomly selecting regions of the input image based on grid sizes $l_i$ and $w_i$, as shown in the formula:

\begin{equation}
\mathbf{m}_{j} = \min\Biggl( 1, \sum_{i=1}^{N} \mathbf{1}_{(l_i, w_i, x)} \Biggr)
\label{eq:6}
\end{equation}
where $\mathbf{1}_{(l_i, w_i, x)}$ is an indicator function used to determine whether a pixel of the input image lies within the $i$-th rectangular region.
We obtain different outputs by randomly selecting regions $T$ times and use them as the feature representations of the SegTrans attack targets.
These regions are dynamically selected during each attack iteration.
Each mask $m$ preserves only the target regions of the attack while setting other regions to zero.
This approach significantly reduces the dependence of the perturbation optimization process on the global features of the input image, effectively minimizing the divergence between the surrogate model and the target model.

\renewcommand{\algorithmicrequire}{\textbf{Input:}} 

\renewcommand{\algorithmicensure}{\textbf{Output:}}

\section{Experiments}

\begin{algorithm}[tb]
\caption{SegTrans}
\label{alg:SegTrans}
\textbf{Input:} Input image $x$, hyper parameters $K$, $T$, $N$ \\
\textbf{Output:} Perturbation $\delta$

\makebox[2.5em][l]{\textbf{0:}} Initialize $\delta \gets 0$ \Comment{Start with no perturbation}

\makebox[2.5em][l]{\textbf{1:}} \textbf{for} $t = 1$ to $K$ \textbf{do} \Comment{Attack iteration}

\makebox[2.5em][l]{\textbf{2:}} \hspace*{1.5em}Divide image $x$ into $N$ random regions with Eq.~\ref{eq:4}

\makebox[2.5em][l]{\textbf{3:}} \hspace*{1.5em}\textbf{for} each of the $N$ random regions \textbf{do}

\makebox[2.5em][l]{\textbf{4:}} \hspace*{3em}Randomly retain some regions of $x$ to form a mask $m$ with Eq.~\ref{eq:6}

\makebox[2.5em][l]{\textbf{5:}} \hspace*{3em}Apply the mask $m$ to the input image $x$, obtaining the enhanced sample $x \odot m$

\makebox[2.5em][l]{\textbf{6:}} \hspace*{1.5em}\textbf{end for}

\makebox[2.5em][l]{\textbf{7:}} \hspace*{1.5em}\textbf{for} $j = 1$ to $T$ \textbf{do} \Comment{Semantic remapping}

\makebox[2.5em][l]{\textbf{8:}} \hspace*{3em}\textbf{for} each enhanced sample \textbf{do}

\makebox[2.5em][l]{\textbf{9:}} \hspace*{4.5em}Recalculate gradients \& update perturbations with Eq.~\ref{eq:5}

\makebox[2.5em][l]{\textbf{10:}} \hspace*{3em}\textbf{end for}

\makebox[2.5em][l]{\textbf{11:}} \hspace*{1.5em}\textbf{end for}

\makebox[2.5em][l]{\textbf{12:}} \textbf{return} $\delta$ \Comment{Final perturbation}
\end{algorithm}

\subsection{Experimental Setup}

\noindent\textbf{Datasets.} 
The adversarial attack experiments in this paper were conducted on the PASCAL VOC2012~\cite{everingham2010pascal} and CITYSCAPES~\cite{cordts2016cityscapes} datasets, both of which are widely used benchmark datasets in semantic segmentation tasks.
The PASCAL VOC2012 dataset is the official dataset used in the PASCAL VOC challenge. This dataset contains $21$ object categories and provides pixel-level annotations for each image. In this study, we use the validation set with a total of $1,449$ images for the adversarial attack tests.
CITYSCAPES dataset includes street scene images collected from $50$ different cities under good weather conditions, with a resolution of $1024\times2048$. 
The dataset covers $19$ categories, including urban street elements like vehicles, pedestrians, and roads.
In this study, we use the validation set with a total of $500$ images for the adversarial attack tests.

\noindent\textbf{Models.} 
We use four segmentation models, including FCN~\cite{long2015fully}, PSPNet~\cite{zhao2017pyramid}, DeepLabV1~\cite{chen2014semantic}, and DeepLabV3+~\cite{xing2020encoder}, along with three backbone networks: MobileNet~\cite{howard2017mobilenets}, ResNet50 and ResNet101~\cite{he2016deep}.

FCN (Fully Convolutional Network)~\cite{long2015fully} is a classic architecture for semantic segmentation tasks in deep learning. Unlike traditional Convolutional Neural Networks (CNNs)\cite{lecun1989backpropagation}, which use upsampling layers after the fully connected layers, FCN replaces the fully connected layers with fully convolutional layers, enabling the network to accept input images of any size and produce dense predictions of the corresponding size. This design allows FCN to handle images of different dimensions and efficiently make pixel-level predictions, which makes it widely used in semantic segmentation.

PSPNet (Pyramid Scene Parsing Network)~\cite{zhao2017pyramid} is a network used for scene parsing. It introduces a Pyramid Pooling Module (PPM) to capture context information at different scales. Through pyramid pooling, PSPNet aggregates global context features of the image while modeling local details, thereby enhancing the understanding of objects at different scales in complex scenes. This multi-scale information fusion strategy leads to significant performance improvements in large-scale semantic segmentation tasks.

DeepLabV1~\cite{chen2014semantic} is another network that achieves significant results in semantic segmentation. 
It uses a standard Convolutional Neural Network (e.g., VGG16 or ResNet) for feature extraction and combines fully connected Conditional Random Fields (CRFs)~\cite{lafferty2001conditional} with the output of the final DCNN\cite{ul2018alexnet} layer to refine segmentation boundaries. 
CRFs model the dependencies between adjacent pixels, helping the network better handle details in the image, particularly in object boundaries and fine regions.

DeepLabV3+~\cite{xing2020encoder} is an optimized version of DeepLabV3, primarily enhancing dilated convolutions and incorporating an encoder-decoder structure to improve semantic segmentation performance. Dilated convolution increases the receptive field of the convolutional kernels, allowing for the capture of a broader context while avoiding an increase in computational cost. The decoder structure helps recover detailed information in the image, especially near segmentation boundaries, thereby enhancing segmentation precision. DeepLabV3+ demonstrates strong capabilities in handling complex scenes and fine-grained segmentation tasks, making it an efficient architecture for semantic segmentation.

\noindent\textbf{Parameter settings.} 
Following~\cite{gu2022segpgd,agnihotri2024cospgd}, we set the upper bound of the perturbation budget to $8/255$.
For our experiments, we set the hyperparameters $N$, $T$, and the side length of the square perturbation regions to $16$, $5$, and $32$, respectively. 



\begin{table*}[htbp]
\setlength{\abovecaptionskip}{4pt}
  \centering
  \caption{The ASR (\%) results of the SegTrans under different settings, with the first and second rows representing different target models and backbone networks.
  Values covered by gray denote the benign mIoU, others denote adversarial mIoU. DeepLabV1, DeepLabV3+, MobileNet, PSPNet, ResNet50, and ResNet101 are abbreviated as DV1, DV3+, M\_Net, P\_Net R\_50, and R\_101, respectively.}
  \scalebox{0.95}{
    \begin{tabular}{ccccccccccccccc}
    \toprule
    \toprule
    \multirow{2}[2]{*}{DATASET} & \multicolumn{2}{c}{\multirow{2}[2]{*}{Model}} & \multicolumn{3}{c}{FCN} & \multicolumn{3}{c}{P\_Net} & \multicolumn{3}{c}{DV1} & \multicolumn{3}{c}{DV3+} \\
    \cmidrule(lr){4-6}\cmidrule(lr){7-9}\cmidrule(lr){10-12}\cmidrule(lr){13-15}  
          & \multicolumn{2}{c}{} & M\_Net & R\_50 & R\_101 & M\_Net & R\_50 & R\_101 & M\_Net & R\_50 & R\_101 & M\_Net & R\_50 & R\_101 \\
    \midrule
    \multicolumn{1}{c}{\multirow{13}[2]{*}{PASCAL VOC}} &       & \cellcolor[rgb]{ .851,  .851,  .851}Benign & \cellcolor[rgb]{ .851,  .851,  .851}52.98  & \cellcolor[rgb]{ .851,  .851,  .851}58.93  & \cellcolor[rgb]{ .851,  .851,  .851}62.46  & \cellcolor[rgb]{ .851,  .851,  .851}66.60  & \cellcolor[rgb]{ .851,  .851,  .851}71.54  & \cellcolor[rgb]{ .851,  .851,  .851}77.97  & \cellcolor[rgb]{ .851,  .851,  .851}53.43  & \cellcolor[rgb]{ .851,  .851,  .851}58.88  & \cellcolor[rgb]{ .851,  .851,  .851}60.54  & \cellcolor[rgb]{ .851,  .851,  .851}70.61  & \cellcolor[rgb]{ .851,  .851,  .851}71.30  & \cellcolor[rgb]{ .851,  .851,  .851}75.62  \\
          & \multirow{3}[0]{*}{FCN} & M\_Net & 52.38  & 38.18  & 39.44  & 65.73  & 43.14  & 35.37  & 52.61  & 39.40  & 38.16  & 69.67  & 40.22  & 35.03  \\
          &       & R\_50 & 42.35  & 57.77  & 55.64  & 50.20  & 69.86  & 64.07  & 42.21  & 57.46  & 53.33  & 52.18  & 68.99  & 62.60  \\
          &       & R\_101 & 41.80  & 51.76  & 61.60  & 47.32  & 62.55  & 76.92  & 40.88  & 51.66  & 59.65  & 51.19  & 58.85  & 73.56  \\
          & \multirow{3}[0]{*}{P\_Net} & M\_Net & 46.29  & 31.60  & 32.12  & 60.74  & 39.59  & 32.11  & 46.60  & 31.98  & 30.32  & 64.18  & 36.89  & 33.68  \\
          &       & R\_50 & 36.04  & 52.16  & 48.01  & 47.85  & 66.36  & 60.84  & 36.03  & 52.01  & 46.70  & 46.93  & 65.26  & 58.09  \\
          &       & R\_101 & 33.03  & 39.93  & 50.11  & 34.75  & 46.24  & 67.93  & 34.90  & 39.77  & 48.73  & 38.63  & 42.44  & 59.55  \\
          & \multirow{3}[0]{*}{DV1} & M\_Net & 51.84  & 39.09  & 39.89  & 65.40  & 44.88  & 35.27  & 52.49  & 39.75  & 40.56  & 69.13  & 42.21  & 38.07  \\
          &       & R\_50 & 41.80  & 57.43  & 54.91  & 49.16  & 69.81  & 62.01  & 41.69  & 57.68  & 52.74  & 51.65  & 68.90  & 61.52  \\
          &       & R\_101 & 40.46  & 49.89  & 61.79  & 47.36  & 60.36  & 77.16  & 39.45  & 49.93  & 59.98  & 48.09  & 56.20  & 73.28  \\
          & \multirow{3}[1]{*}{DV3+} & M\_Net & 45.01  & 26.63  & 27.80  & 57.87  & 31.31  & 31.87  & 44.36  & 27.63  & 26.72  & 63.03  & 29.83  & 24.95  \\
          &       & R\_50 & 36.53  & 51.72  & 48.03  & 47.82  & 64.34  & 60.16  & 36.18  & 51.48  & 46.75  & 51.81  & 65.09  & 58.83  \\
          &       & R\_101 & 37.44  & 47.10  & 56.23  & 46.86  & 60.21  & 71.39  & 37.59  & 47.26  & 54.59  & 48.42  & 58.27  & 70.51  \\
\cmidrule{2-15}    \multicolumn{1}{c}{\multirow{13}[2]{*}{CITY\newline{}SCAPES}} &       & \cellcolor[rgb]{ .851,  .851,  .851}Benign & \cellcolor[rgb]{ .851,  .851,  .851}56.36  & \cellcolor[rgb]{ .851,  .851,  .851}58.66  & \cellcolor[rgb]{ .851,  .851,  .851}60.63  & \cellcolor[rgb]{ .851,  .851,  .851}56.92  & \cellcolor[rgb]{ .851,  .851,  .851}60.58  & \cellcolor[rgb]{ .851,  .851,  .851}61.37  & \cellcolor[rgb]{ .851,  .851,  .851}56.49  & \cellcolor[rgb]{ .851,  .851,  .851}57.27  & \cellcolor[rgb]{ .851,  .851,  .851}59.30  & \cellcolor[rgb]{ .851,  .851,  .851}61.94  & \cellcolor[rgb]{ .851,  .851,  .851}62.67  & \cellcolor[rgb]{ .851,  .851,  .851}64.12  \\
          & \multirow{3}[0]{*}{FCN} & M\_Net & 56.03  & 51.31  & 51.38  & 55.43  & 49.53  & 45.65  & 56.11  & 49.91  & 50.33  & 60.52  & 51.28  & 47.83  \\
          &       & R\_50 & 44.21  & 58.19  & 56.68  & 39.03  & 58.74  & 52.69  & 44.62  & 56.72  & 55.24  & 44.06  & 60.60  & 60.59  \\
          &       & R\_101 & 38.98  & 52.99  & 60.03  & 33.25  & 51.00  & 59.71  & 39.10  & 51.25  & 58.73  & 37.80  & 51.46  & 61.75  \\
          & \multirow{3}[0]{*}{P\_Net} & M\_Net & 55.98  & 50.07  & 48.64  & 55.73  & 47.99  & 42.70  & 56.20  & 48.65  & 47.02  & 60.86  & 49.84  & 45.09  \\
          &       & R\_50 & 40.34  & 58.19  & 56.72  & 35.57  & 59.12  & 53.01  & 40.66  & 56.69  & 55.41  & 41.12  & 60.92  & 54.40  \\
          &       & R\_101 & 41.63  & 54.61  & 60.21  & 36.00  & 51.73  & 60.21  & 41.17  & 52.95  & 58.84  & 38.53  & 52.38  & 63.10  \\
          & \multirow{3}[0]{*}{DV1} & M\_Net & 55.98  & 51.79  & 51.76  & 55.36  & 50.53  & 46.27  & 56.13  & 50.09  & 50.74  & 60.38  & 51.84  & 49.13  \\
          &       & R\_50 & 45.61  & 58.12  & 56.79  & 40.07  & 58.55  & 52.85  & 45.81  & 56.78  & 55.51  & 45.43  & 60.59  & 54.10  \\
          &       & R\_101 & 40.19  & 53.85  & 60.04  & 34.60  & 51.71  & 59.74  & 39.96  & 52.05  & 58.68  & 38.97  & 52.20  & 61.88  \\
          & \multirow{3}[1]{*}{DV3+} & M\_Net & 55.95  & 50.36  & 48.08  & 55.66  & 48.27  & 41.76  & 56.12  & 48.08  & 46.85  & 60.93  & 51.99  & 43.92  \\
          &       & R\_50 & 39.68  & 57.89  & 54.89  & 34.57  & 58.52  & 49.53  & 40.01  & 56.43  & 53.03  & 40.94  & 61.15  & 50.45  \\
          &       & R\_101 & 41.75  & 55.13  & 60.17  & 35.90  & 53.08  & 60.12  & 41.47  & 53.69  & 58.81  & 38.65  & 53.93  & 63.09  \\
    \bottomrule
    \bottomrule
    \end{tabular}%
    }
  \label{tab:main}%
\end{table*}%

\begin{table*}[h]
  \centering
    \caption{The ASR(\%) results of comparison study. The bolded values represent the highest attack success rate under the current setting. DV3+ with R\_101 refers to the DeepLabV3+ model with a ResNet101 backbone, and P\_Net with R\_50 refers to the PSPNet model with a ResNet50 backbone. Target model and Surrogate model are abbreviated as T Model and S Model, PSPNet, DeepLabV1, DeepLabV3+ are abbreviated as PSP, DV1 and DV3+, respectively.}
  \scalebox{1.2}{
    \begin{tabular}{ccccccccccccc}
    \toprule
    \toprule
    T Model & \multicolumn{6}{c}{DV3+ with R\_101}    & \multicolumn{6}{c}{P\_Net with R\_50} \\
\cmidrule(lr){2-7} \cmidrule(lr){8-13}\   Dataset & \multicolumn{3}{c}{PASCAL VOC} & \multicolumn{3}{c}{CITYSCAPES} & \multicolumn{3}{c}{PASCAL VOC} & \multicolumn{3}{c}{CITYSCAPES} \\
\cmidrule(lr){2-4} \cmidrule(lr){5-7} \cmidrule(lr){8-10} \cmidrule(lr){11-13}\    S Model & \multicolumn{1}{l}{FCN} & \multicolumn{1}{l}{PSP} & \multicolumn{1}{l}{DV1} & \multicolumn{1}{l}{FCN} & \multicolumn{1}{l}{PSP} & \multicolumn{1}{l}{DV1} & \multicolumn{1}{l}{FCN} & \multicolumn{1}{l}{DV1} & \multicolumn{1}{l}{DV3+} & \multicolumn{1}{l}{FCN} & \multicolumn{1}{l}{DV1} & \multicolumn{1}{l}{DV3+} \\
\cmidrule(lr){2-13}    
    PGD   & 37.93 & 39.02 & 36.26 & 44.36 & 38.53 & 43.52 & 23.72 & 24.71 & 15.03 & 18.66 & 20.34 & 15.91 \\
    S-PGD & 41.34 & 40.14 & 37.99 & 45.90  & 42.89 & 46.23 & 22.68 & 23.54 & 16.11 & 20.68 & 21.04 & 18.91 \\
    T-S-PGD & 39.20  & 44.16 & 35.83 & 46.80  & 44.02 & 48.05 & 26.69 & 24.56 & 15.58 & 22.58 & 22.26 & 19.30 \\
    C-PGD & 39.15 & 41.61 & 43.36 & 47.63 & 44.44 & 50.63 & 28.82 & 21.67 & 17.49 & 20.36 & 20.25 & 18.39 \\
    MI-FGSM & 51.10 & 49.68 & 49.90 & 50.51 & 48.35 & 50.82 & 37.06 & 37.86 & 27.28 & 33.70 & 34.19 & 30.92 \\
    D-Scaling & 40.13 & 40.10 & 39.45 & 49.92 & 45.97 & 49.87 & 34.98 & 34.45 & 22.02 & 30.76 & 30.04 & 26.28 \\
    A-FGSM & 40.13 & 40.10 & 39.45 & 54.24 & 49.90 & 53.72 & 36.92 & 38.85 & 29.95 & 35.21 & 35.62 & 32.42 \\
    EBAD  & 49.91 & 48.24 & 44.04 & 53.87 & 51.54 & 50.61 & 38.73 & 36.95 & 25.86 & 38.33 & 36.14 & 42.24 \\
    Ours  & \textbf{62.60} & \textbf{58.09} & \textbf{61.52} & \textbf{60.59} & \textbf{54.40} & \textbf{54.10} & \textbf{43.14} & \textbf{44.88} & 
    
    \textbf{31.31} & \textbf{49.53} & \textbf{50.53} & \textbf{48.27} \\
    \bottomrule
    \bottomrule
    \end{tabular}%
    }
  \label{tab:comparison}%
\end{table*}%

\begin{figure*}[!t]
    \centering
    \includegraphics[scale=0.6]{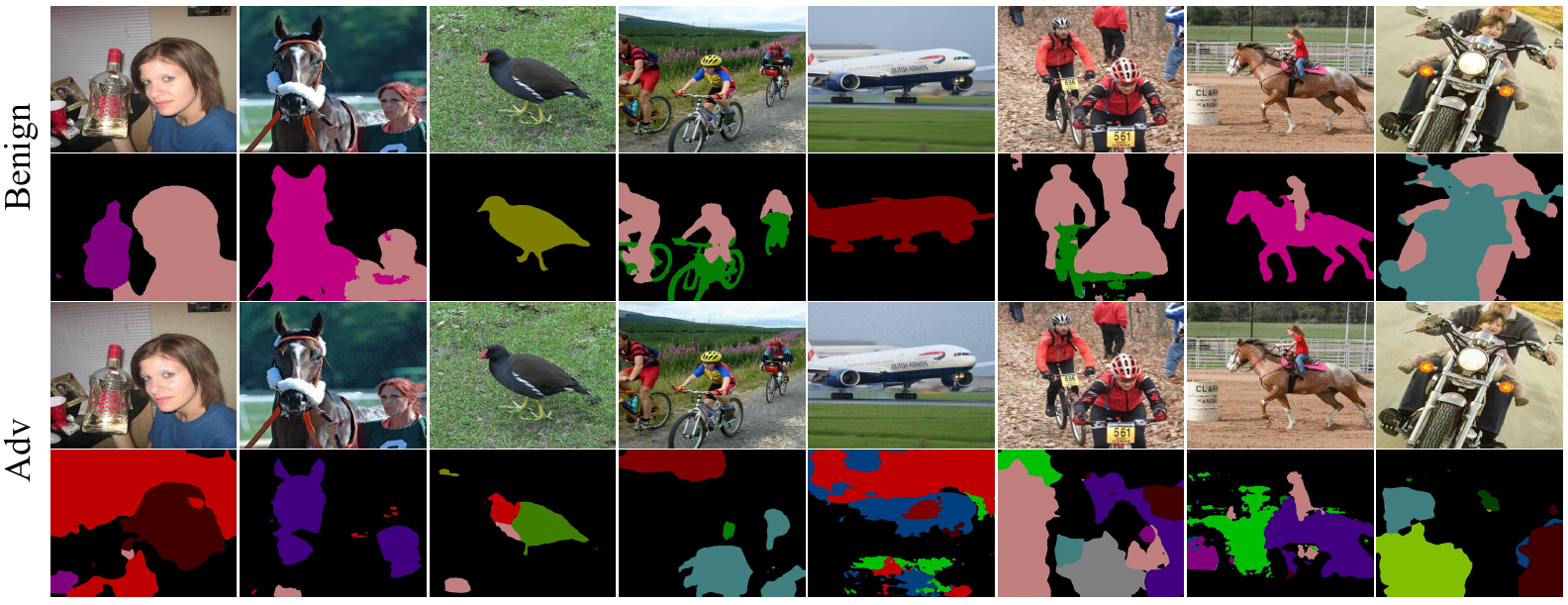}
    \caption{The visualization results of adversarial examples by SegTrans attack}
    \label{fig:attack}
\end{figure*}

\begin{figure*}[http]
    \centering
    \includegraphics[scale=0.6]{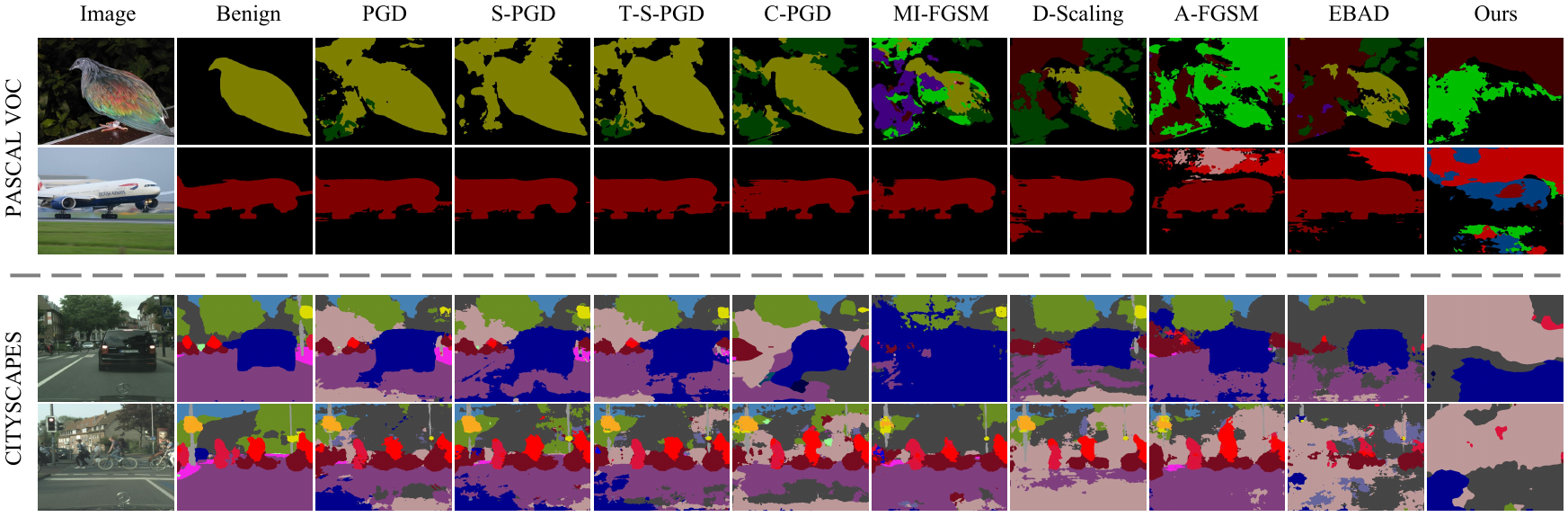}

    \caption{Visualization of comparison results. The upper part shows the PASCAL VOC dataset, and the lower part shows the CITYSCAPES dataset. 
    The first column displays the original samples, with each of the remaining columns representing the segmentation results of different adversarial attack methods.}

    \label{fig:comparison}

\end{figure*}

\noindent\textbf{Evaluation metrics.} 
We use \textit{Mean Intersection over Union} (mIoU)~\cite{long2015fully} as the metric to evaluate segmentation accuracy. 
mIoU is a commonly used evaluation method in semantic segmentation tasks to measure the model's segmentation performance across different pixel categories.
It is calculated by determining the \textit{Intersection over Union} (IoU) for all pixel categories in a sample and averaging the IoUs across all categories to obtain the final evaluation result.
Specifically, IoU is the ratio of the intersection area between the predicted region and the ground truth region to the union of both areas. 
The formula for calculating IoU for each sample is:
\begin{equation}
    \text{IoU} = \frac{\text{Predicted Region} \cap \text{Ground Truth Region}}{\text{Predicted Region} \cup \text{Ground Truth Region}}
\nonumber
\end{equation}

The mIoU is the average of the IoUs for different pixel categories across all samples, reflecting the model's overall performance in segmentation tasks. 
A higher mIoU indicates better segmentation performance across categories, especially in tasks with class imbalance or fine-grained segmentation. 
Therefore, mIoU is commonly used as an important metric to evaluate the performance of semantic segmentation models.

In addition to mIoU, we also use the \textit{Attack Success Rate} (ASR) to evaluate the effectiveness of adversarial attacks.
ASR measures the performance of the semantic segmentation model under adversarial attacks, representing the difference between the mIoU values of benign and adversarial samples when the model is attacked. 
A higher ASR indicates a higher attack success rate and poorer model robustness.
The formula is as follows:
\begin{equation}
    \text{ASR} = \text{mIoU}_{\text{Benign}} - \text{mIoU}_{\text{Adversarial}}
    \nonumber
\end{equation}
where the \(\text{mIoU}_{\text{benign}}\) refers to the mIoU of samples when no adversarial perturbations are applied, and \(\text{mIoU}_{\text{adversarial}}\) refers to the mIoU of samples after the model undergoes adversarial attacks.

\noindent\textbf{Experimental hardware details.} 
We conduct experiments on a machine with two NVIDIA A100-SXM4 GPUs, two Intel(R) Xeon(R) Gold 6132 CPUs and 314GB RAM. 

\begin{figure}[http]   
  \centering
      \subcaptionbox{PASCAL VOC}{\includegraphics[width=0.23\textwidth]{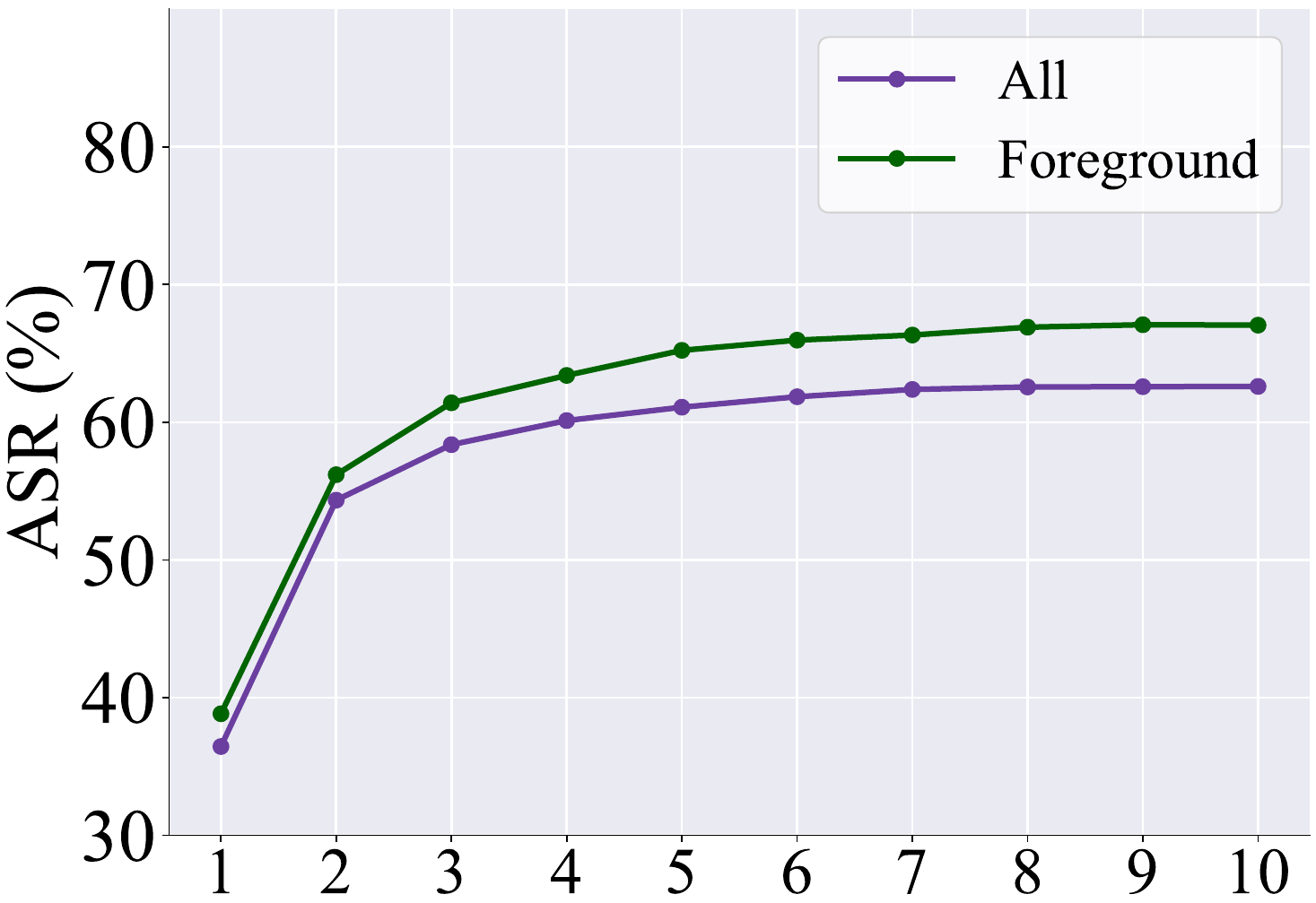}}
      \subcaptionbox{CITYSCAPES}{\includegraphics[width=0.23\textwidth]{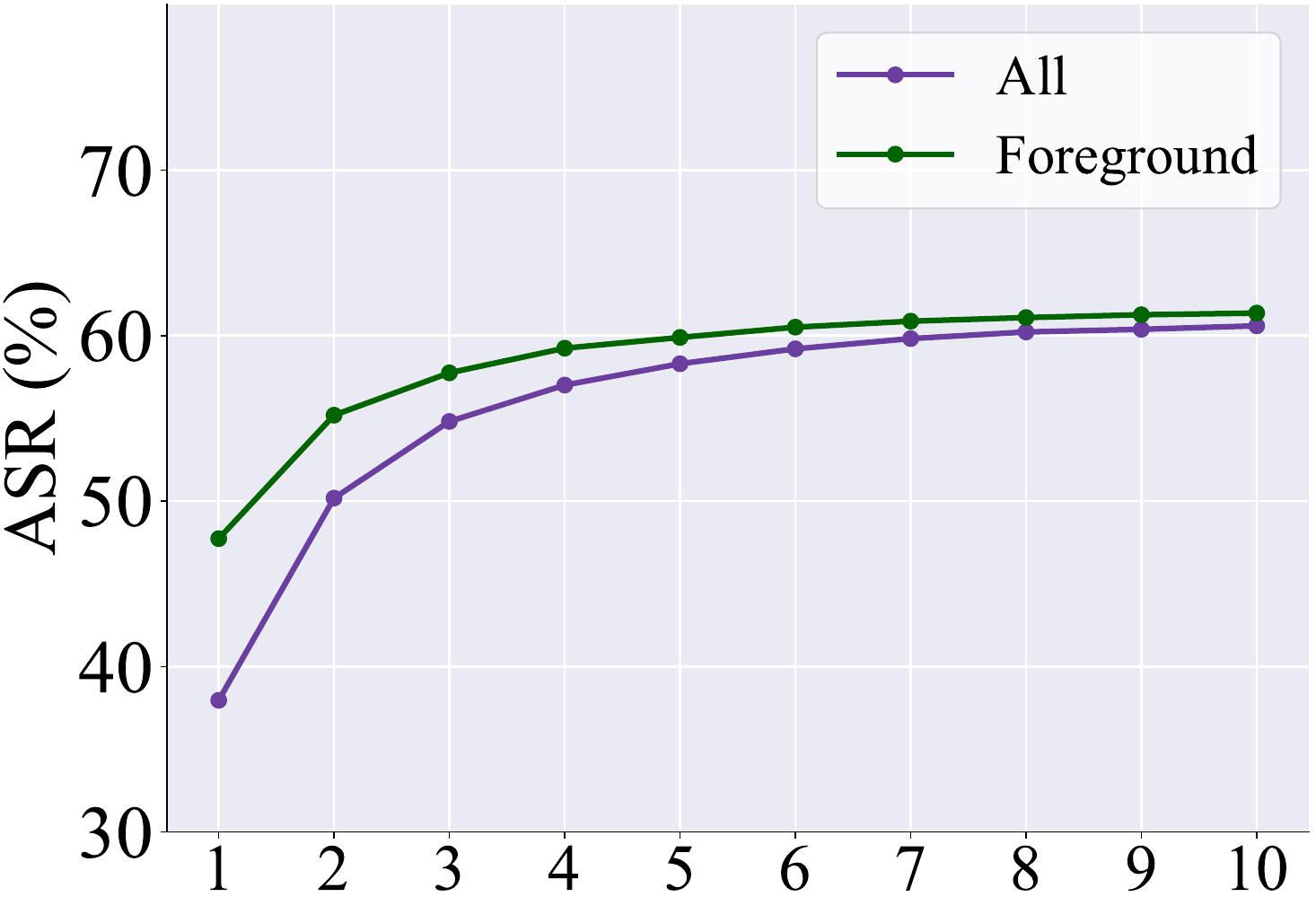}}
      \caption{Subfigures (a) and (b) respectively show the impact on attack performance after removing background information on different datasets, with the horizontal axis representing the number of attack iterations.}
       \label{fig:fore}
        \vspace{-0.4cm}
\end{figure}

\subsection{Attack Performance}

In this section, we comprehensively evaluate the effectiveness of SegTrans. 
We conduct attack experiments on four segmentation models that utilize three different backbone networks, evaluated across two datasets.
We evaluate the mean Intersection over Union (mIoU) of both benign and adversarial examples under each experimental setup and calculate the corresponding ASR.

According to the experimental results in~\cref{tab:main}, SegTrans can effectively deceive these segmentation models, achieving an ASR of over 50\% across 288 different experimental settings.
It is worth noting that the choice of surrogate model has little impact on attack effectiveness, as all selected surrogate models are able to consistently generate adversarial examples with strong attack performance.
In addition to the above experimental results, we also present the visualization of adversarial example outputs, as shown in~\cref{fig:attack}.
It can be seen that our attack method almost destroys all semantic information in the benign image, further validating the powerful attack capability of SegTrans.

To further verify the impact of tight coupling on attack success rate, we add a comparative experiment on attack success rates with and without background samples, as shown in~\cref{fig:fore} (a) and (b). 
We evaluate the attack success rates of SegTrans after each iteration on the PASCAL VOC and CITYSCAPES datasets. 
The results show that the attack success rate for samples without background (represented by the green line) is significantly higher than for samples with background. 
This confirms the observations we made in~\cref{fig:tight}.
Additionally, we investigate the impact of the black regions outside the selected area on perturbation generation. 
Specifically, we replace the enhanced image with a pure black image to guide the perturbation optimization process. 
It is important to note that since all pixel values in a pure black image are zero, it does not contain any meaningful semantic information.
The experimental setup is consistent with Section 4.4.
According to the experimental results, the ASR under this setting is only 2.76\%, which is close to the effect of adding random Gaussian noise to the original image (2.53\%). 
This result further confirms the effectiveness of the Multi-region Perturbation Activation strategy and Semantic Remapping strategy.

\begin{figure}[http]   
  \centering
      \subcaptionbox{Convergence nalysis}{\includegraphics[width=0.23\textwidth]{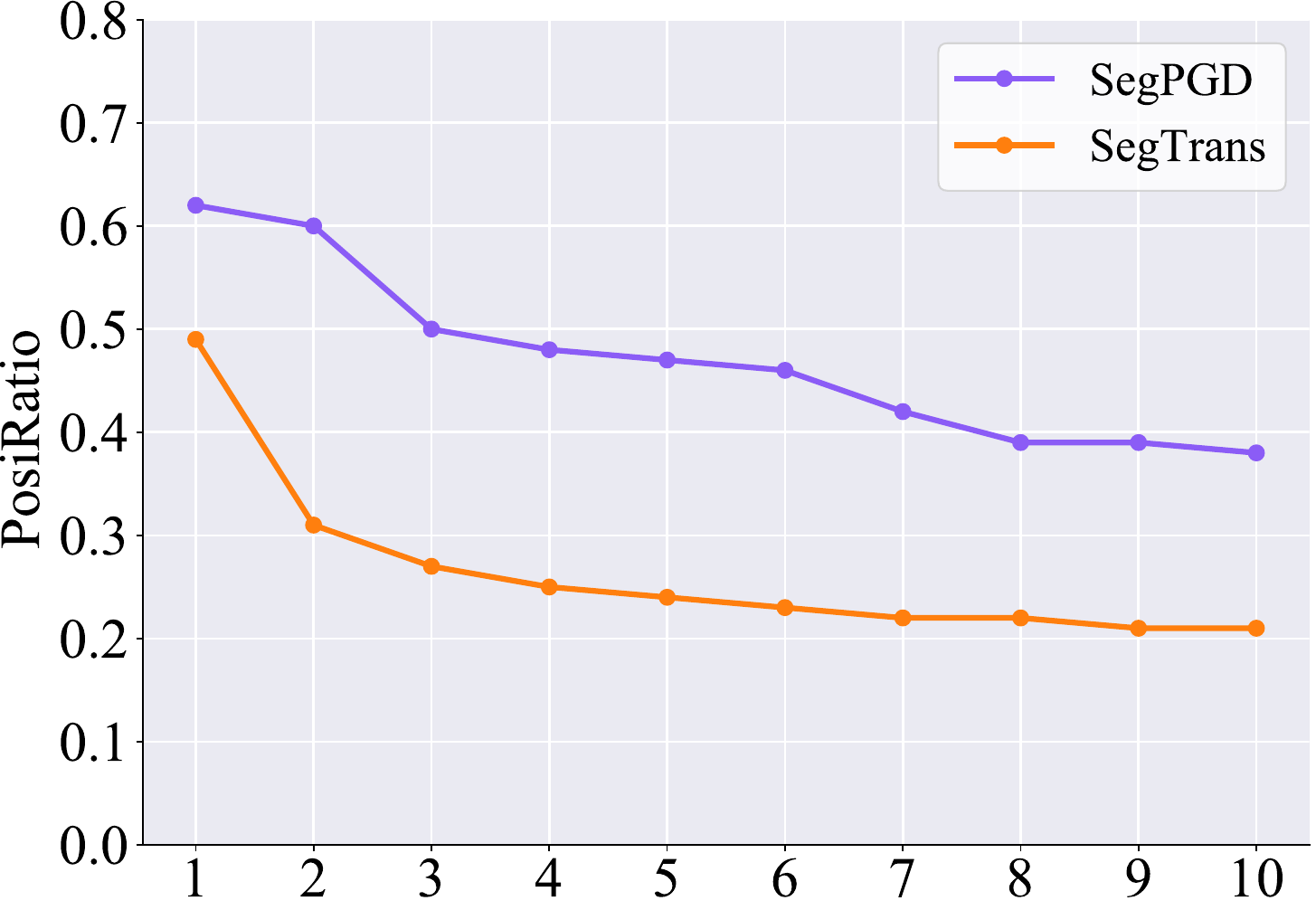}}
      \subcaptionbox{Random seed}{\includegraphics[width=0.23\textwidth]{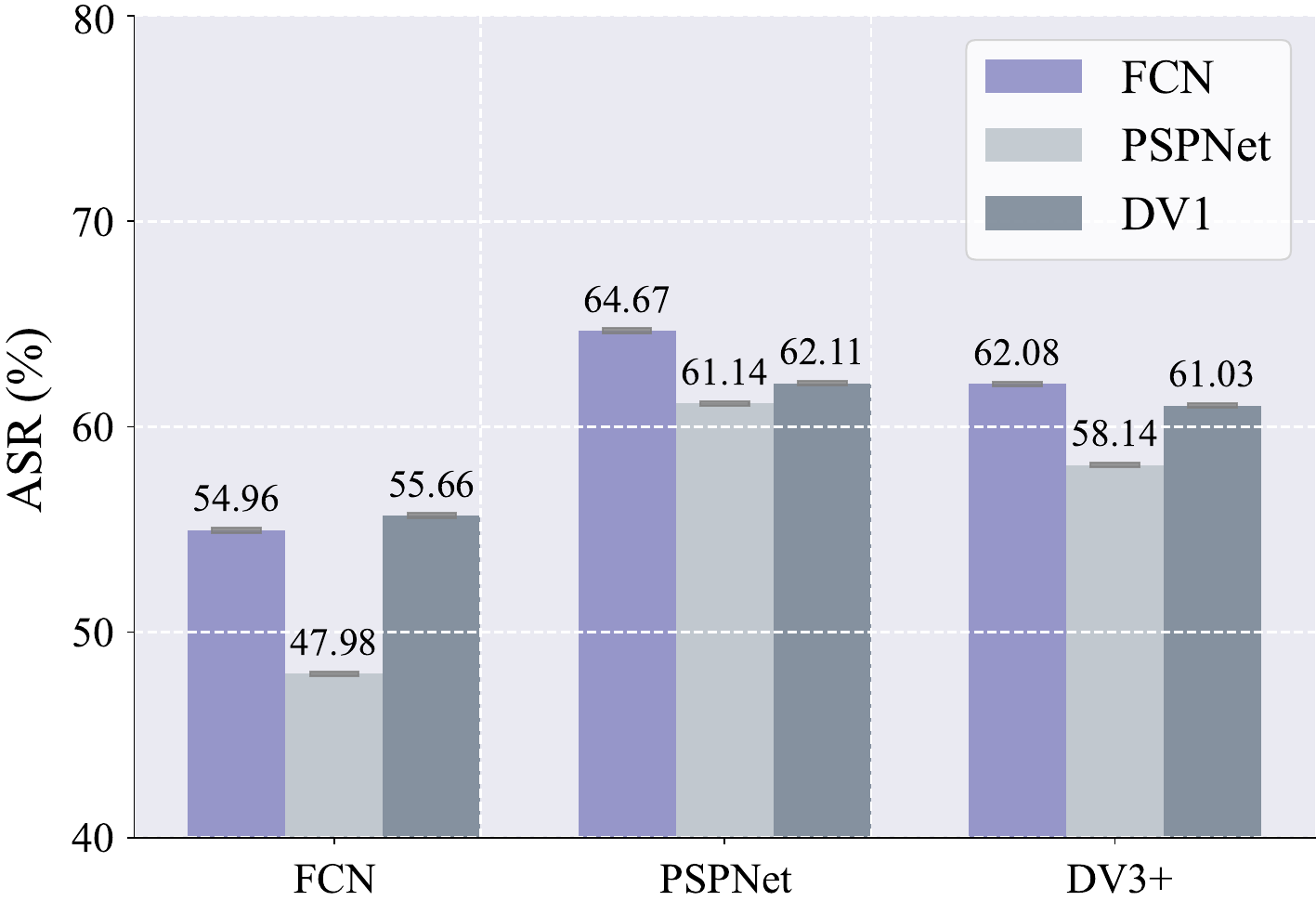}}

    
      \caption{Subfigures (a) and (b) illustrate the convergence analysis and the impact of different random seeds on attack performance, respectively.}
       \label{fig:add}
\end{figure}

\subsection{Comparison Study}
In this section, we compare SegTrans with popular adversarial attack methods, including adversarial attacks designed for classification models such as PGD~\cite{PGD}, MI-FGSM~\cite{dong2018boosting}, D-Scaling~\cite{gu2021adversarial}, and Advanced-FGSM~\cite{maag2024uncertainty}, as well as adversarial attacks targeting segmentation models, including SegPGD~\cite{gu2022segpgd}, CosPGD~\cite{agnihotri2024cospgd}, TranSegPGD~\cite{jia2023transegpgd} and the \textit{state-of-the-art} (SOTA) method EBAD~\cite{cai2023ensemble}.
We select DeepLabV3+ and PSPNet as the target models and use ResNet101 and ResNet50 as their respective backbones.
The experimental datasets are PASCAL VOC and CITYSCAPES.
To ensure a fair comparison, we keep the attack parameters consistent across all methods.
The results in~\cref{tab:comparison} show that our method outperforms all other methods by a significant margin.
Notably, SegTrans consistently outperforms the SOTA method EBAD, with an average ASR improvement of 8.55\%.
The comparative attack visualizations in~\cref{fig:comparison} further demonstrate the significant advantages of our method.

In~\cref{tab:time}, we also provide a comparison of the computational efficiency of the ten methods. 
Specifically, we choose DeepLabV3+ as the target model with ResNet101 as its backbone and FCN as the surrogate model with ResNet50 as its backbone.
The results show that SegTrans’s computational efficiency is close to that of PGD, SegPGD, TranSegPGD, and CosPGD, with a difference of no more than 6\% compared to the most computationally efficient PGD. 
Notably, SegTrans significantly outperforms EBAD in attack performance, while achieving over 200\% of EBAD’s computational efficiency.
This demonstrates that our method does not introduce additional computational overhead while achieving excellent transfer success rates.
Additionally, we present a detailed comparison of the stealthiness of 10 adversarial attack methods in~\cref{tab:psnr}, using PSNR~\cite{hore2010image} as the evaluation metric. 
SegTrans achieves an average PSNR of 30.57, which is slightly lower than the overall average of 36.43 across the 8 methods.
Notably, according to the attack results in~\cref{tab:comparison} and the visualizations in~\cref{fig:comparison}, SegTrans achieves significantly better performance than SOTA methods, with an average increase of 8.55\% in attack success rate. 
This demonstrates that our method strikes a good balance between attack effectiveness and visual stealthiness.

To further demonstrate the superiority of SegTrans, we compare the convergence efficiency of SegTrans and SegPGD, recording the PosiRatio values for 10 iterations. 
Specifically, we follow the experimental setup and definitions from SegPGD~\cite{gu2022segpgd}, where PosiRatio = 1 - MisRatio, with MisRatio defined as the ratio of misclassified pixels to all input pixels.
The target model is DeepLabV3+ with ResNet101 as the backbone, and the surrogate model is FCN with ResNet50 as the backbone.
The results in ~\cref{fig:add}(a) show that our approach clearly outperforms SegPGD in convergence efficiency, further confirming the superior attack performance of SegTrans.

\begin{table*}[!t]
\setlength{\abovecaptionskip}{4pt}
  \centering
  \caption{Time consumption analysis of different attack methods.}
  \resizebox{0.8\textwidth}{!}{%
    \begin{tabular}{ccccccccccc}
    \toprule
    \toprule
          & Methods & PGD   & S-PGD & T-S-PGD & C-PGD  & MI-FGSM & D-Scaling & A-FGSM & EBAD  & Ours \\
\cmidrule{2-11}    \multirow{2}[2]{*}{PASCAL VOC} & ASR(↑) & 37.93 & 41.34 & 39.20  & 39.15 & 51.10  & 40.13 & 40.13 & 49.91 & \textbf{62.60} \\
          & Samples/s(↑) & 2.36/s & 2.38/s & 2.27/s & 2.32/s & 2.41/s & 2.28/s & 2.30/s & 1.00/s & 1.93/s \\
\cmidrule{2-11}    \multirow{2}[2]{*}{CITYSCAPES} & ASR(↑) & 44.36 & 45.90  & 46.80  & 47.63 & 50.51 & 49.92 & 54.24 & 53.87 & \textbf{60.59} \\
          & Samples/s(↑) & 2.86/s & 2.88/s & 2.61/s & 2.92/s & 2.80/s & 3.04/s & 3.05/s & 1.05/s & 2.19/s \\
    \bottomrule
    \bottomrule
    \end{tabular}%
    }
  \label{tab:time}%
\end{table*}%

\begin{figure*}[!t]   
\setlength{\abovecaptionskip}{4pt}
  \centering
       \subcaptionbox{Grid counts number}
       {\includegraphics[width=0.194\textwidth]{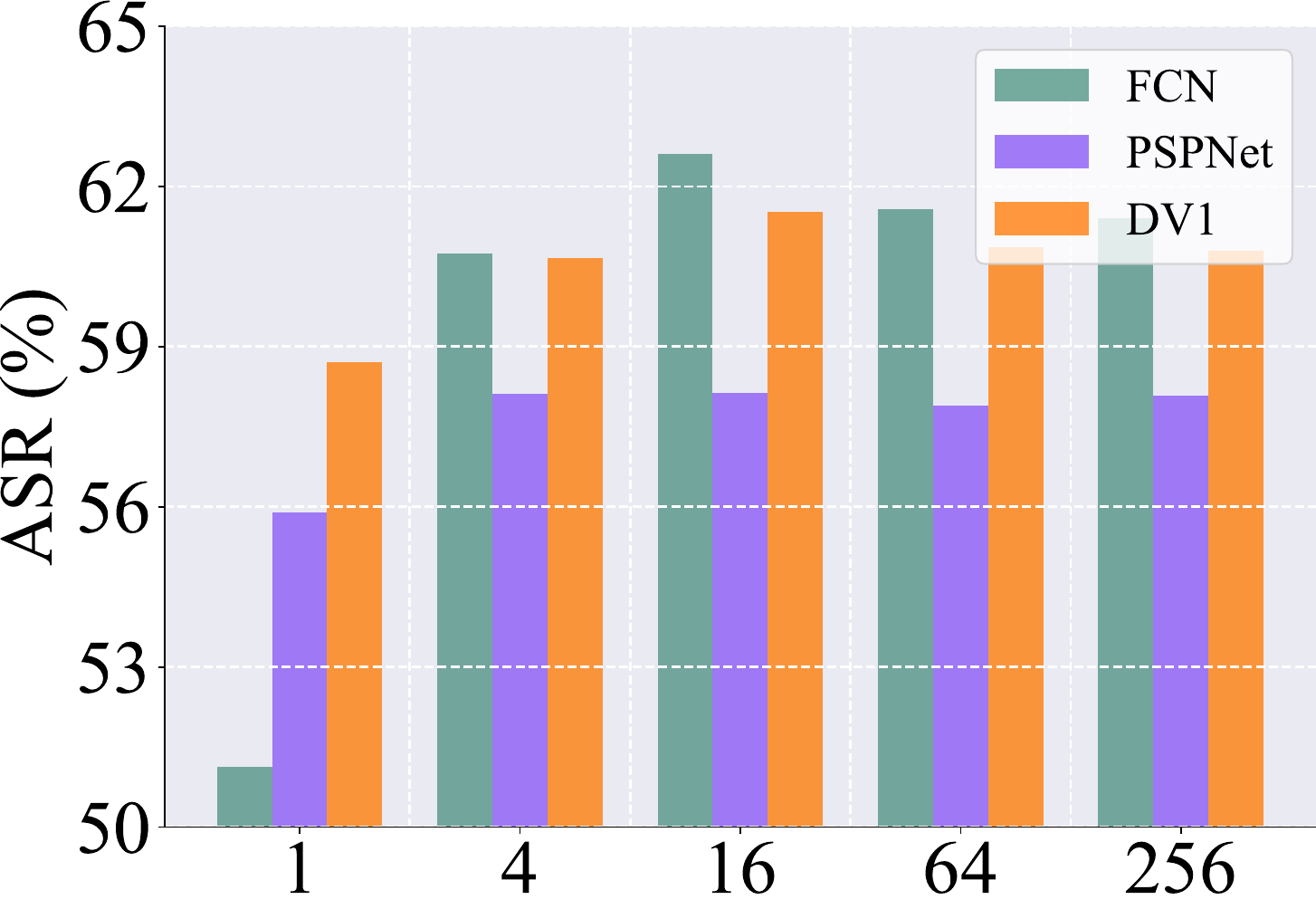}}
    \subcaptionbox{Length}{\includegraphics[width=0.194\textwidth]{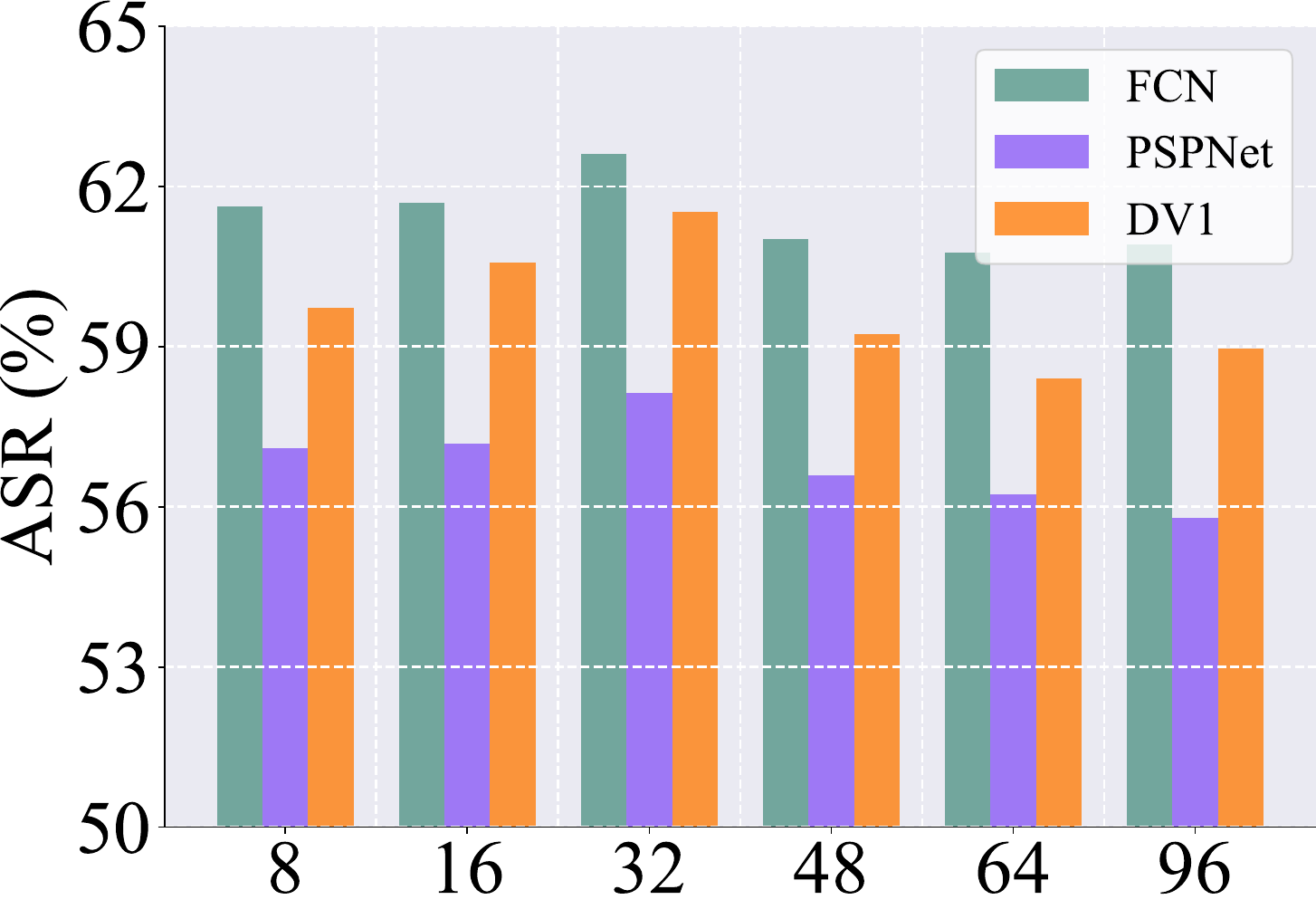}}
    \subcaptionbox{Perturbation budget}{\includegraphics[width=0.194\textwidth]{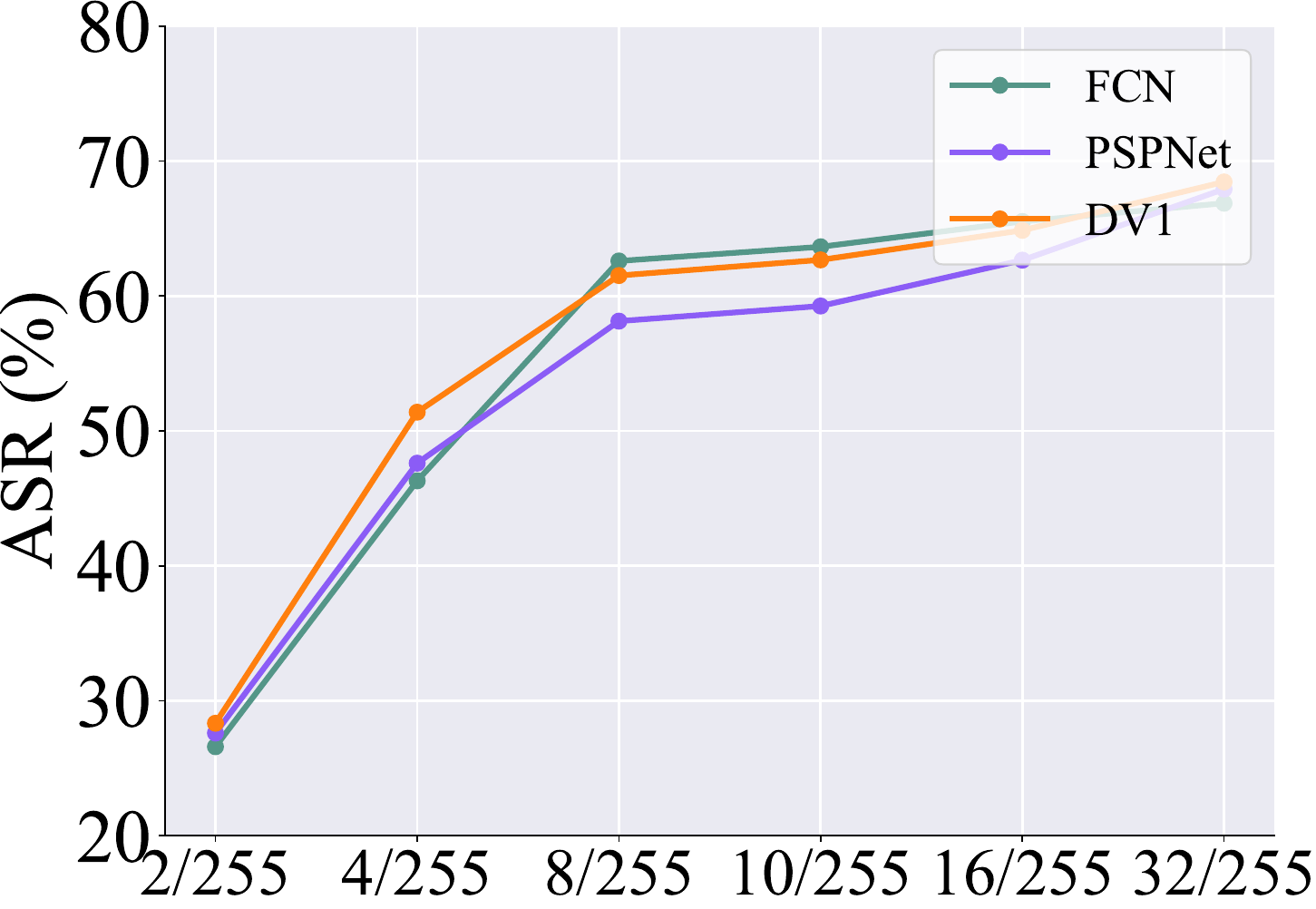}}
     \subcaptionbox{Remapping iterations}{\includegraphics[width=0.194\textwidth]{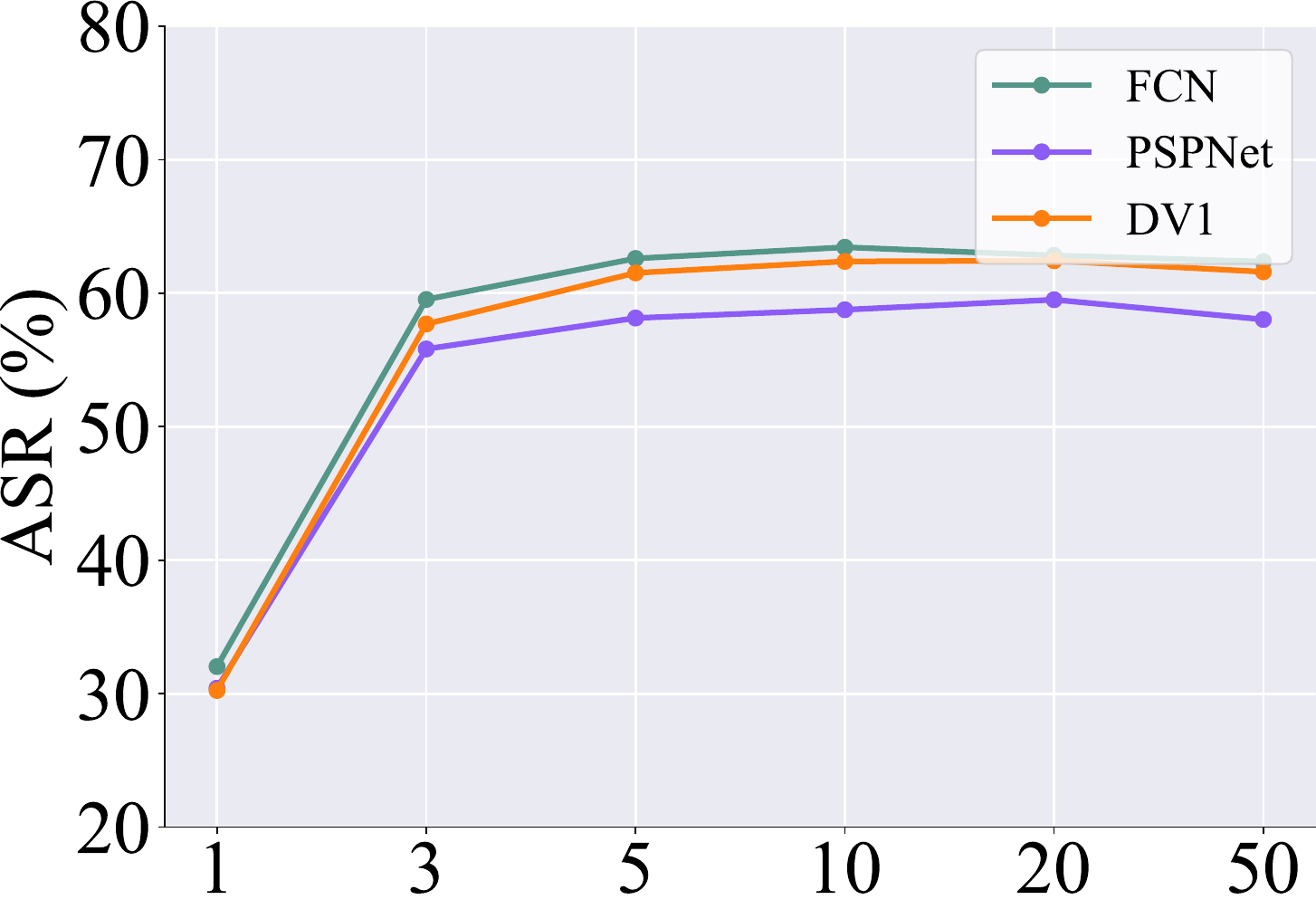}}
       \subcaptionbox{Epoch}{\includegraphics[width=0.194\textwidth]{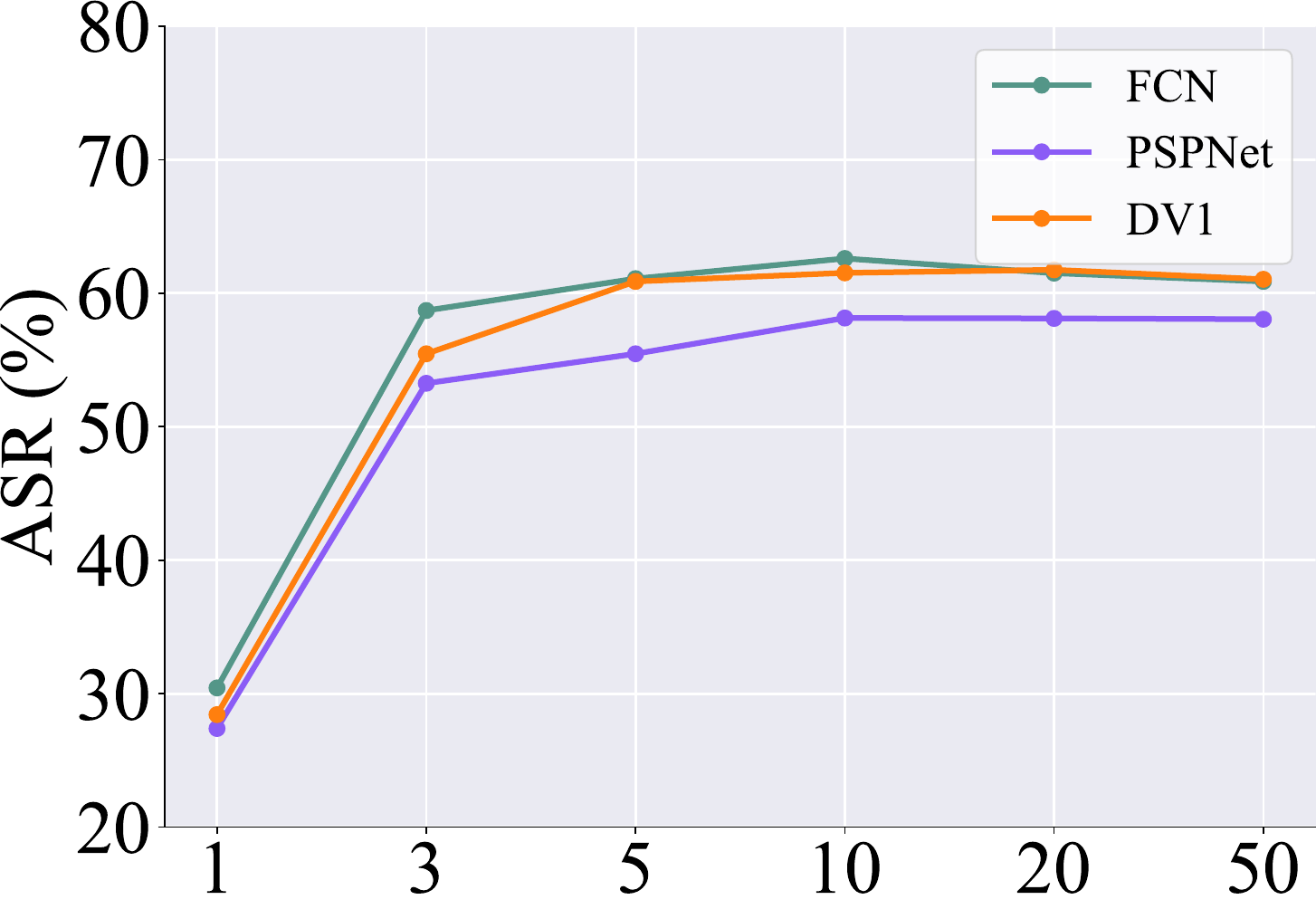}}
      \caption{The ASR (\%) results of ablation study. Subfigures (a) - (e) investigate the effect of different numbers of the grid count, side length of random region, attack strengths, semantic remapping iterations, and attack iterations on SegTrans, respectively.}
       \label{fig:ablation_results}
\end{figure*}



 


    

\begin{table}[htbp]
  \centering
  \caption{The PSNR results of different adversarial attack methods.}
  \resizebox{0.46\textwidth}{!}{%
    \begin{tabular}{ccccccccc}
    \toprule
    \toprule
    \multirow{2}[4]{*}{Setting} & \multicolumn{4}{c}{PASCAL VOC} & \multicolumn{4}{c}{CITYSCAPES} \\
\cmidrule(lr){2-5}  \cmidrule(lr){6-9}            & FCN   & PSPNet & DV1   & DV3+  & FCN   & PSPNet & DV1   & DV3+ \\
\cmidrule(lr){2-9}     PGD   & 37.65  & 37.66  & 37.70  & 37.52  & 37.97  & 38.01  & 37.99  & 38.03  \\
    S-PGD & 37.66  & 37.66  & 37.71  & 37.51  & 37.94  & 38.00  & 37.96  & 38.01  \\
    T-S-PGD & 37.75  & 37.84  & 37.78  & 37.77  & 37.96  & 38.01  & 37.99  & 38.03  \\
    C-PGD & 37.75  & 37.84  & 37.78  & 37.77  & 38.00  & 38.12  & 38.03  & 38.15  \\
    MI-FGSM & 32.22  & 32.21  & 32.23  & 32.20  & 32.02  & 32.18  & 32.05  & 32.22  \\
    D-Scaling & 37.88  & 37.90  & 37.91  & 37.84  & 38.02  & 38.08  & 38.03  & 38.09  \\
    A-FGSM & 31.99  & 32.06  & 31.99  & 32.04  & 32.00  & 32.04  & 31.98  & 31.96  \\
    EBAD  & 37.70  & 37.62  & 37.70  & 37.62  & 38.00  & 38.01  & 38.07  & 37.99  \\
    Ours  & 30.58  & 30.52  & 30.58  & 30.49  & 30.62  & 30.57  & 30.62  & 30.58  \\
    \bottomrule
    \bottomrule
    \end{tabular}%
    }
  \label{tab:psnr}%
\end{table}%

\subsection{Ablation Study}
In this section, we explore the effect of different grid counts, random perturbation region side length, attack strengths, semantic remapping iterations, and attack iterations on SegTrans. The surrogate models are FCN, PSPNet, and DeepLabV1, with ResNet50 as the backbone network. The target model is DeepLabV3+, using ResNet101 as the backbone network. The dataset is PASCAL VOC.

\noindent\textbf{The effect of the number of the grid counts.} 
We study the effect of the number of grids in the proposed Multi-region Perturbation Activation strategy.
We conduct experiments with varying numbers of grid counts, ranging from $1$ to $256$.
The results shown in~\cref{fig:ablation_results}(a) indicate that the attack performance stabilizes when the number of grids is $4$, and reaches its best when the number of grids is $16$.

\noindent\textbf{The effect of side length of random region.} 
We study the effect of the size of each random perturbation region. 
We conduct experiments with varying numbers of side lengths of square regions ranging from 8 to 96.
The results shown in~\cref{fig:ablation_results}(b) indicate that attack performance is optimized when the side length of the square is 32, followed by a downward trend. 
This can be attributed to the fact that excessively large square regions lead to masks containing redundant information, thereby impacting the attack efficacy.

\noindent\textbf{The effect of perturbation budget.} 
As shown in~\cref{fig:ablation_results}(c), we evaluate SegTrans's attack performance with $\epsilon$ ranging from 2/255 to 32/255. 
With the increase in $\epsilon$ , there is a corresponding enhancement in attack performance.
Notably, our attack still maintains high efficacy at the 4/255 setting, with the average attack success rate exceeding 48.43\%.

\noindent\textbf{The effect of semantic remapping iterations.} 
We study the effect of the number of semantic remapping on attack performance. 
We conduct experiments with varying numbers
of semantic remapping iterations, ranging from 1 to 50. 
The results shown in~\cref{fig:ablation_results}(d) indicate that the attack performance stabilizes after the number of iterations reaches 5.
Notably, the attack performance at 5 iterations (the default setting) is very close to that at 10 iterations, with a difference in mIoU of less than 1\%.

\noindent\textbf{The effect of attack iterations.} 
We study the effect of the attack iterations on the performance of adversarial examples. 
We conduct experiments with varying numbers
of attack iterations, ranging from 1 to 50. 
The results shown in~\cref{fig:ablation_results}(e) indicate that the attack performance stabilizes after 5 iterations, with the optimal attack performance achieved at 10 iterations (the default setting).

\subsection{Stability Analysis}
We analyze the impact of different random seeds on the attack performance of SegTrans on the PASCAL VOC dataset. 
Specifically, we select FCN, PSPNet, and DeepLabV1 as surrogate models, and FCN, PSPNet, and DeepLabV3+ as target models, aiming to assess the stability of the attack performance under different random seeds.
The experimental results in ~\cref{fig:add}(b) show that although variations in random seeds may cause some performance fluctuations, our attack method exhibits a small standard error across different seeds. 
This indicates that the attack performance remains stable under different initialization conditions, further validating the effectiveness of our approach.



 

\subsection{Adversarial Defense}

We discuss whether SegTrans is resistant to adversarial defense methods that include adversarial training~\cite{zhang2024atzsl,yang2023noise}, model pruning~\cite{xiang2024adapmtl,lu2024generic}, and corruption~\cite{cohen2019certified}.

\begin{figure}[!t]   
  \centering
      \subcaptionbox{AT-FGSM}{\includegraphics[width=0.23\textwidth]{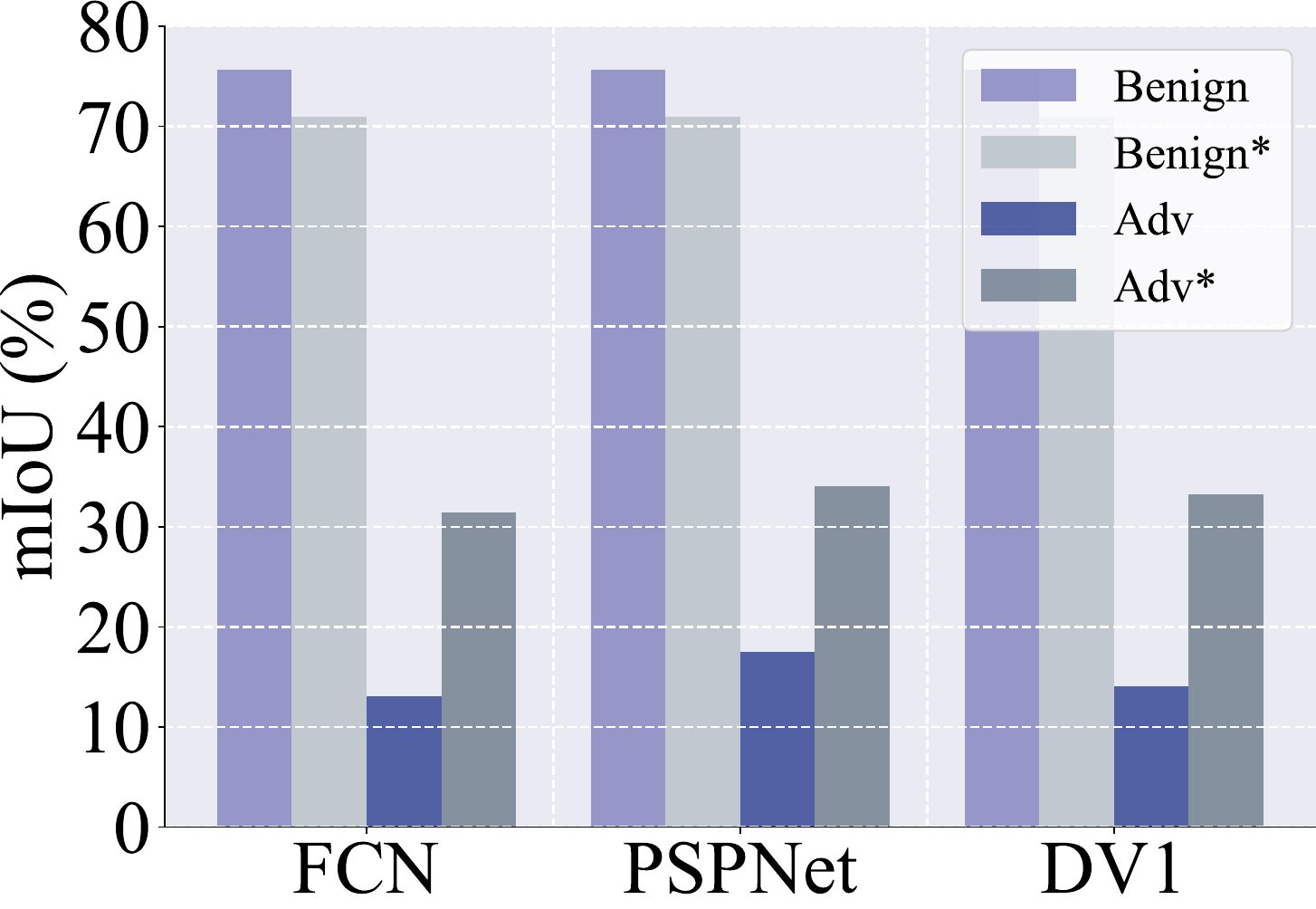}}
      \subcaptionbox{AT-SegPGD}{\includegraphics[width=0.23\textwidth]{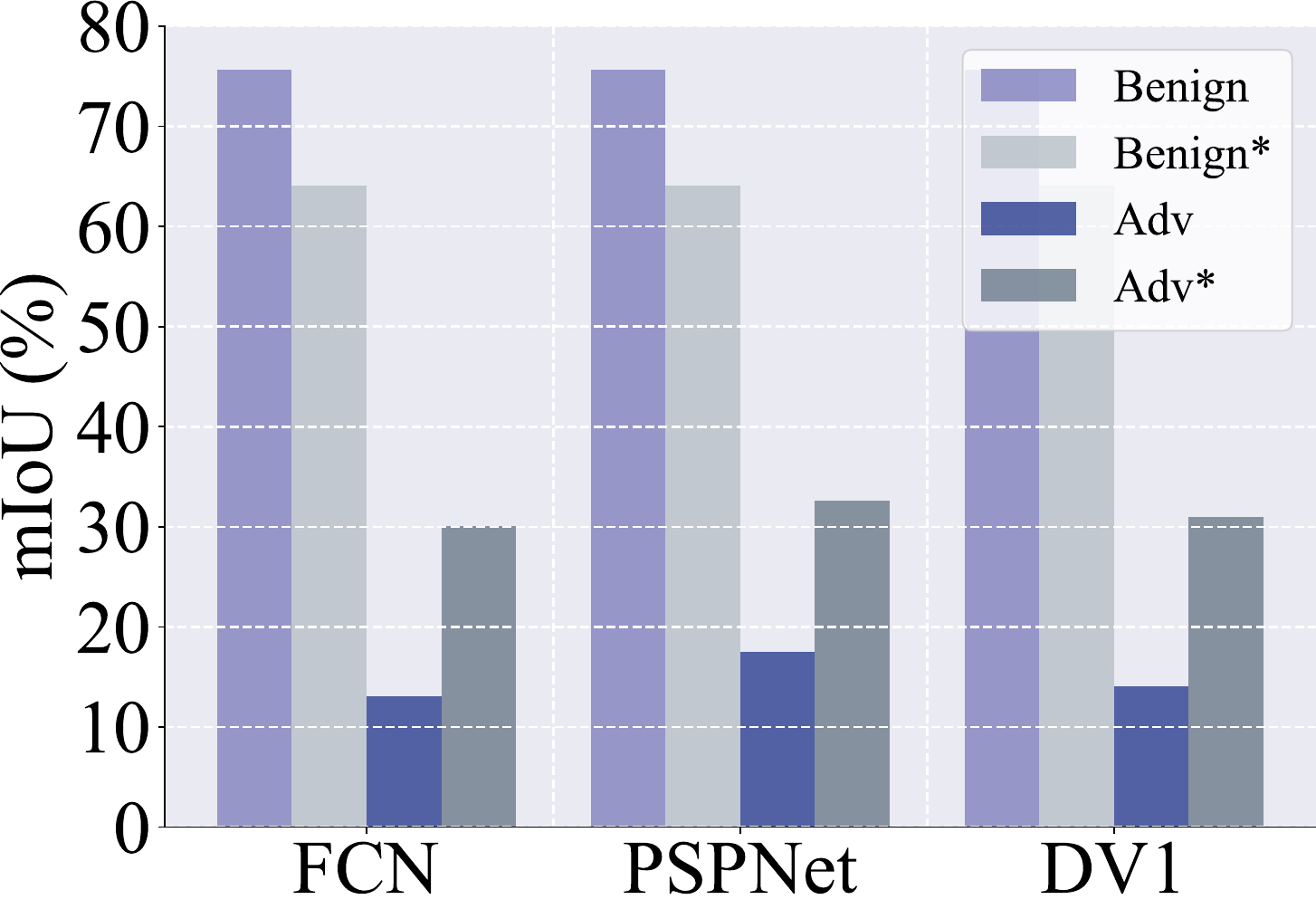}}
      \subcaptionbox{Pruning}{\includegraphics[width=0.23\textwidth]{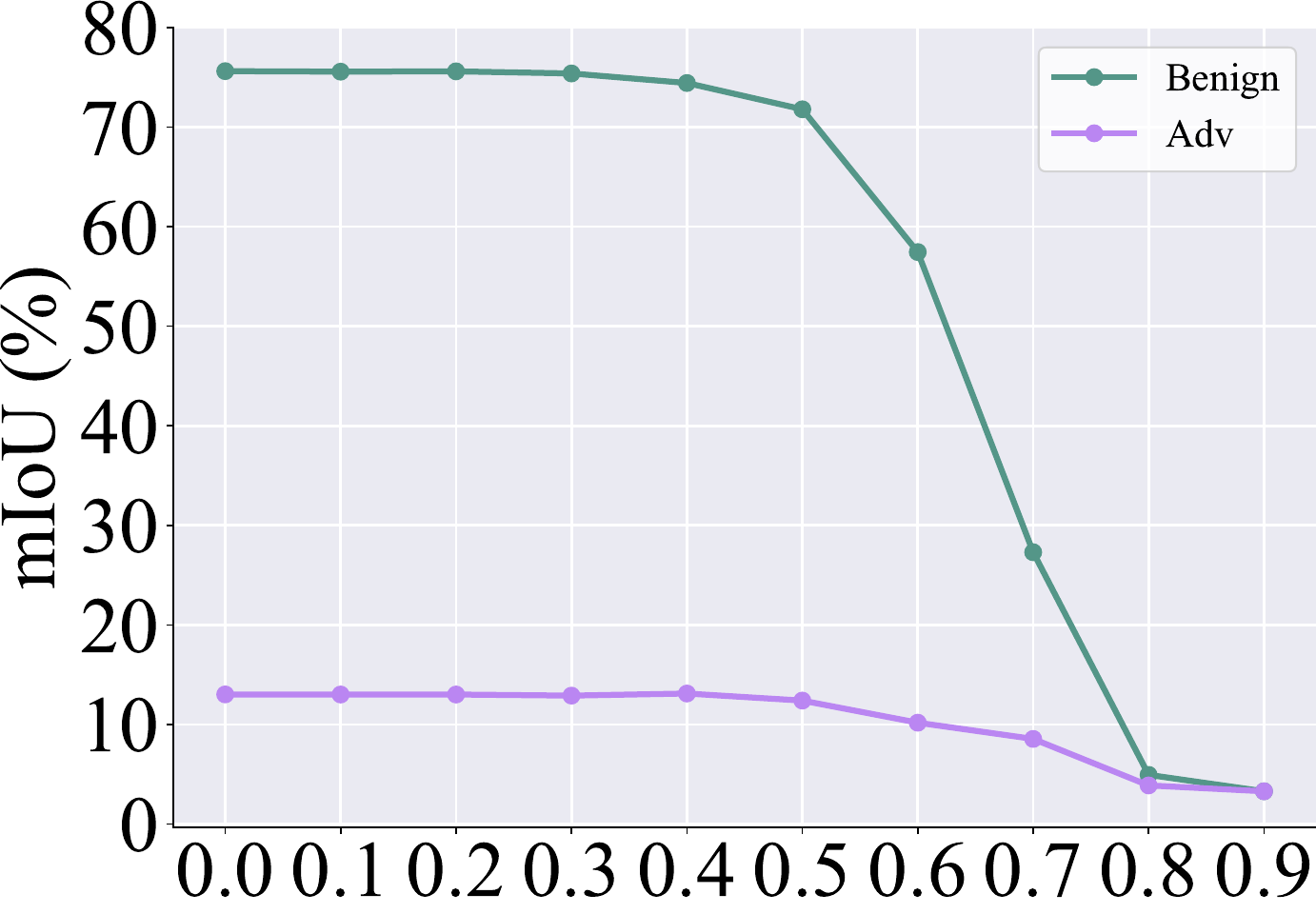}}
      \subcaptionbox{Corruption}{\includegraphics[width=0.23\textwidth]{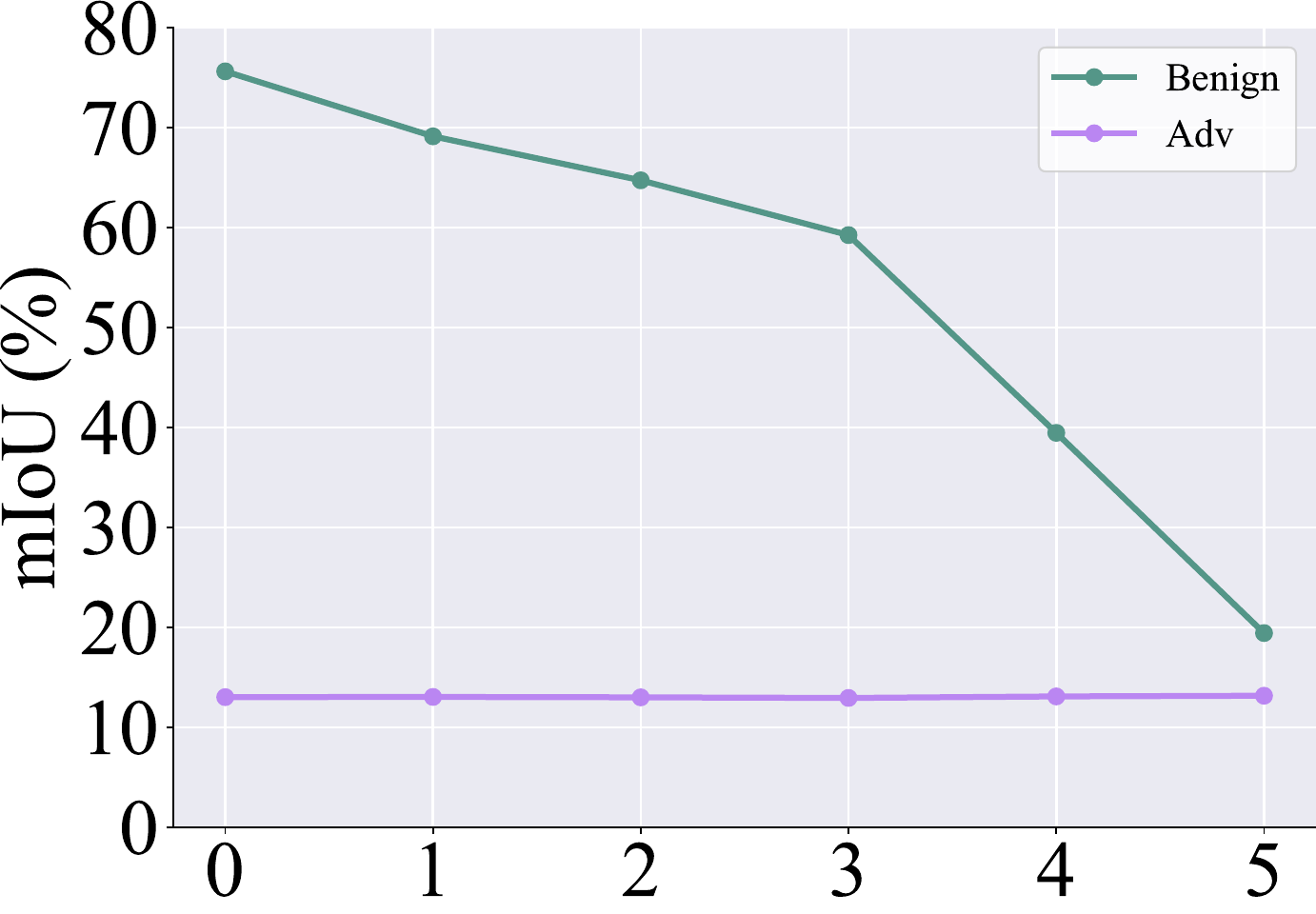}}
    
      \caption{The attack performance of SegTrans against different defense methods on the PASCAL VOC dataset.
      Subfigures (a) and (b) examine the adversarial training method, while (c) and (d) examine the Pruning and Corruption methods, respectively.}
       \label{fig:defense}
        \vspace{-0.4cm}
\end{figure}

\noindent\textbf{Adversarial training.}
We apply adversarial training to the DeepLabV3+ model, with the experimental setup consistent with Section 4.4. 
The attack methods used are FGSM\cite{goodfellow2014explaining} and SegPGD.
As shown in~\cref{fig:defense}(a)-(b), the results marked with an asterisk (“*”) represent the performance after adversarial training.
The results indicate that, although adversarial training slightly weakens the effectiveness of our attack, the mIoU of the adversarial examples remains below 40\%, demonstrating that SegTrans still maintains strong transferability even when facing defense models.

\noindent\textbf{Model pruning.} 
We perform model pruning on DeepLabV3+ with pruning rates ranging from 0 to 0.9, as outlined in Section 4.3. Results in~\cref{fig:defense}(c) show that as the pruning rate increases, the mIoU of benign samples gradually decreases, while the mIoU of adversarial samples remains nearly unchanged. Even at a pruning rate of 0.7, where model performance degrades significantly, the mIoU for adversarial examples remains 8.55\%, indicating that pruning is ineffective against SegTrans.

\noindent\textbf{Data corruption.} 
We evaluate the effectiveness of corruption as a defense mechanism against SegTrans attacks, where the corruption method involves adding Gaussian noise to adversarial examples, with corruption levels ranging from 0 to 5. 
The experimental setup is consistent with Section 4.3.
As shown in~\cref{fig:defense}(d), as the corruption level increases, the mIoU of benign samples gradually decreases, while the mIoU of adversarial examples remains almost unchanged. 
This indicates that SegTrans can effectively resist the defense of corruption.

\section{Limitations and Future Work}
\label{sec:limitations}
In this work, we primarily focus on transferable adversarial attacks targeting semantic segmentation models. 
A key question is whether adversarial training using SegTrans can enhance the robustness of semantic segmentation models against a broader range of adversarial attacks. 
We plan to explore this issue further in future work, as it could have a significant impact on improving the robustness of semantic segmentation models and advancing the field. 
Another key issue is that we have not yet extended our approach to classification~\cite{singh2021metamed} or object detection~\cite{feng2024security} models due to structural differences between these models. 
However, we believe that the proposed multi-region perturbation activation and semantic remapping strategies offer a novel perspective with practical application potential for adversarial attack research.
Future work can explore how to apply these strategies to tasks beyond semantic segmentation. 
Additionally, in the adversarial defense experiments, our research lacks sufficient theoretical analysis and does not explain why the proposed attacks can effectively bypass existing defense mechanisms, which limits the deep understanding of adversarial defense strategies. 
Therefore, we will focus on addressing this theoretical gap in future research to promote the design and optimization of defense mechanisms.
\section{Conclusions}
\label{sec:Conclusion}
In this paper, we propose SegTrans, a transfer attack method for segmentation models.
To address the issue of tight coupling phenomenon in segmentation tasks, we introduce a multi-region perturbation activation strategy that disrupts the global coherence of input features by randomly retaining portions of the sample's semantic information.
Subsequently, we design an innovative semantic remapping method to mitigate the impact of feature fixation on the attack. 
By perturbing the critical object features in the input samples, our method effectively reduces the feature representation discrepancy between the surrogate model and the target model, thereby successfully deceiving the segmentation models. 
Our extensive experiments on FCN, PSPNet, DeepLabV1, and DeepLabV3+ across three backbones and two datasets demonstrate the strong attack performance and transferability of SegTrans.

\section*{Acknowledgements}
Shengshan Hu's work is supported by the National Natural Science Foundation of China under Grant No.62372196.
Minghui Li's work is supported by the National Natural Science Foundation of China under Grant No.62572206.

\bibliographystyle{IEEEtran}
\bibliography{tmm2025}

\end{document}